\documentclass[conference]{IEEEtran}
\usepackage{times}

\usepackage[numbers]{natbib}
\usepackage{multicol}
\usepackage{hyperref}

\usepackage{booktabs}
\usepackage{multirow}

\usepackage{amsmath}
\usepackage{amssymb}

\usepackage[table, dvipsnames]{xcolor} 
\usepackage{colortbl}

\definecolor{RoyalBlue}{RGB}{65,105,225}
\definecolor{lightblue}{RGB}{235,244,253}

\usepackage{pifont}
\newcommand{\cmark}{\ding{51}}
\newcommand{\xmark}{\ding{55}}

\usepackage{graphicx}
\usepackage{xspace}

\usepackage{cuted}
\usepackage{caption}

\usepackage{fontawesome5}
\usepackage{marvosym}

\usepackage{tabularx} 

\begin{document}

\title{Emerging Extrinsic Dexterity in Cluttered Scenes via Dynamics-aware Policy Learning}

\author{
    Yixin Zheng$^{*1,2,3}$, Jiangran Lyu$^{*3,4,\dagger}$, Yifan Zhang$^3$, Jiayi Chen$^{3,4}$, Mi Yan$^{3,4}$, Yuntian Deng$^{5}$\\ 
    Xuesong Shi$^{3}$, Xiaoguang Zhao$^1$, Yizhou Wang$^{4}$, Zhizheng Zhang$^{2,3, \text{\Letter}}$, He Wang$^{2,3,4, \text{\Letter}}$
    \vspace{0.1cm} \\
    \small $^1$Institute of Automation, Chinese Academy of Sciences \quad $^2$Beijing Academy of Artificial Intelligence \\ 
    $^3$Galbot \quad $^4$Peking University \quad $^5$Shanghai Jiao Tong University \vspace{0.1cm} \\
    $^*$ Equal contribution \quad $^\dagger$ Project Lead \quad \text{\Letter} Corresponding authors
    \thanks{\vspace{-0.1cm}\textbf{Project Page:} \color{RoyalBlue}\textbf{https://pku-epic.github.io/DAPL}}
}

\IEEEoverridecommandlockouts

\maketitle

\begin{strip}\centering
\vspace{-1.6cm}
\includegraphics[width=1\linewidth]{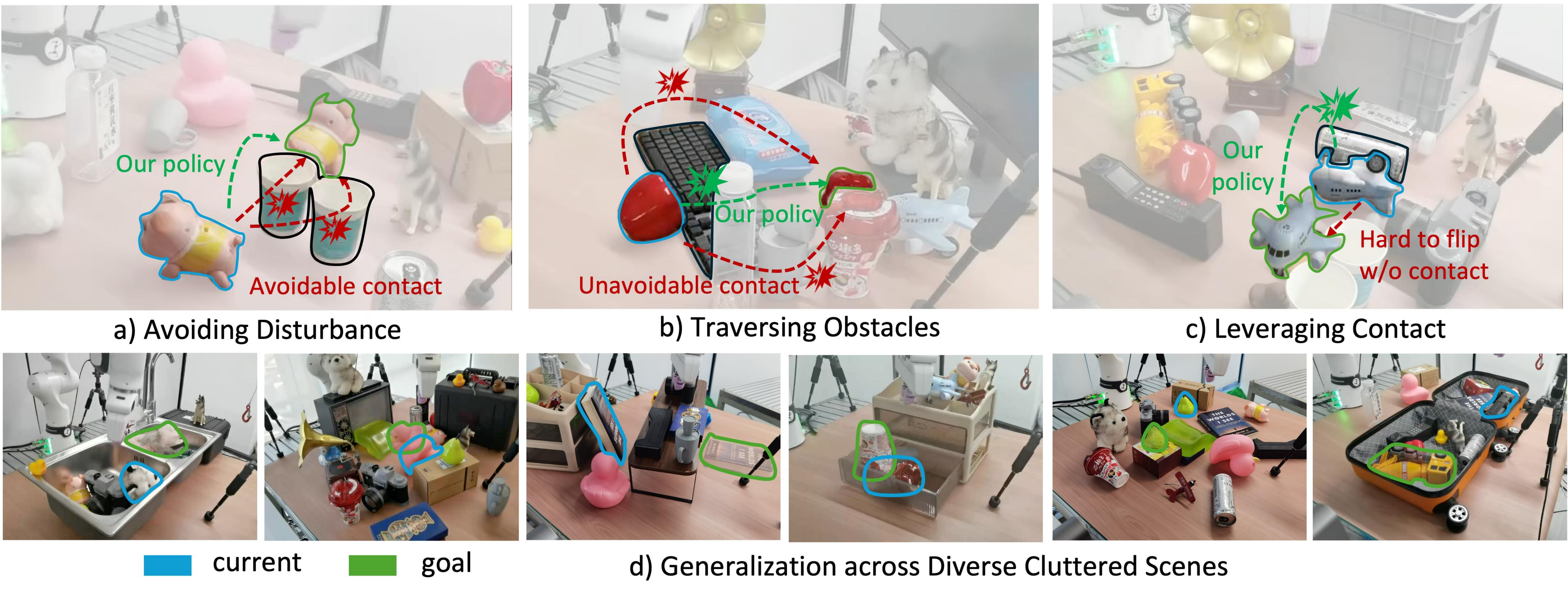}
\vspace{-0.5cm}
\captionof{figure}{Given the current object state (blue), the policy aims to rearrange objects to the goal state (green).
To achieve this in cluttered scenes, the robot policy leverages extrinsic dexterity selectively.
(a) The policy avoids unnecessary contacts to minimize disturbance when free-space motion suffices.
(b) When contact is unavoidable, it robustly traverses obstacles under contact-rich dynamics.
(c) It actively leverages environmental contacts to accomplish otherwise difficult manipulations (e.g., flipping objects).
(d) The approach generalizes across diverse cluttered environments.
}

\label{first_figure} 
\vspace{-.2cm}
\end{strip}

\begin{abstract}
Extrinsic dexterity leverages environmental contact to overcome the limitations of prehensile manipulation. However, achieving such dexterity in cluttered scenes remains challenging and underexplored, as it requires selectively exploiting contact among multiple interacting objects with inherently coupled dynamics. Existing approaches lack explicit modeling of such complex dynamics and therefore fall short in non-prehensile manipulation in cluttered environments, which in turn limits their practical applicability in real-world environments. In this paper, we introduce a Dynamics-Aware Policy Learning  (DAPL) framework that can facilitate policy learning with a learned representation of contact-induced object dynamics in cluttered environments. This representation is learned through explicit world modeling and used to condition reinforcement learning, enabling extrinsic dexterity to emerge without hand-crafted heuristics or complex reward engineering.
We evaluate our approach in both simulation and real-world. Our method outperforms prehensile manipulation, human teleoperation, and prior representation-based policies by over 25\% in success rate on unseen simulated cluttered scenes with varying densities. 
The real-world success rate reaches around 50\% across 10 cluttered scenes, while a practical grocery deployment further demonstrates robust sim-to-real transfer and applicability.


\end{abstract}

\IEEEpeerreviewmaketitle

\section{Introduction}
Robotic manipulation in cluttered scenes demands strategies beyond prehensile actions. Objects are often tightly packed and occluded in cluttered environments, where incidental and coupled contacts are pervasive, making reliable grasps difficult and collision-free execution highly constrained. In such cases, effective manipulation often relies on extrinsic dexterity—the ability to selectively leverage or avoid contacts with the surrounding environment through non-prehensile actions such as pushing, sliding, or toppling—to achieve task goals that grasping alone cannot.

Existing non-prehensile approaches either rely on highly specific designs or simplify away the complexity of contact-rich interactions. Model-based planning and hand-designed primitives~\cite{cheng2023enhancing,oller2024tactile,wu2024one} do not scale beyond narrowly defined settings, while reinforcement learning methods~\cite{kobashi2025leveraging,yang2024learning} are typically confined to simplified contact scenarios. Even recent geometry-centric representation learning approaches, such as CORN\cite{cho2024corn} and UniCORN\cite{cho2025hierarchical}, remain brittle when faced with dense clutter and multiple interacting contacts. All these works lack explicit modeling of complex dynamics and therefore fall short in manipulation in cluttered environments.

As discussed above, achieving extrinsic dexterity in cluttered scenes remains underexplored and challenging, as it requires selectively exploiting beneficial contacts while avoiding disruptive ones. For example, a policy may need to bypass light objects (Fig.\ref{first_figure}(a)), navigate through unavoidable obstacles (Fig.\ref{first_figure}(b)), or leverage stable neighbors to flip a target object (Fig.\ref{first_figure}(c))—tasks that cannot be solved by grasping alone. 
In such cluttered scenarios, success is not determined by geometry alone, but by how objects respond once contact occurs—whether they slide, topple, or transfer motion to surrounding clutter. These complex interactions depend on underlying object dynamics and cannot be reliably inferred from static observations or sparse rewards. This motivates us to decouple dynamics representation learning from task-specific control and to equip the policy with an explicit, learned representation that models contact-induced object motion.

In this paper, we introduce a \textbf{Dynamics-Aware Policy Learning (DAPL)}  framework for more effective non-prehensile manipulation in cluttered scenes. It facilitates the emergence of extrinsic dexterity through contact-induced scene dynamics representation. Specifically, our approach learns this representation via a physical world model trained to predict object motion under interaction, capturing how contacts among multiple objects influence future dynamics. The learned dynamics representation is then used to condition reinforcement learning, enabling policies to reason about the consequences of contact and to selectively exploit environmental interactions.
We further introduce curriculum learning into the DAPL framework, which uses policy rollout trajectories to iteratively refine the dynamics representation. As a result, extrinsic dexterity emerges naturally rather than being hand-designed or shaped by complex rewards.

We evaluate our approach extensively in both simulation and the real-world. To enable systematic training and benchmarking, we introduce \textbf{Clutter6D}, a simulation environment for 6D object rearrangement in cluttered scenes with varying densities. In simulation, DAPL significantly outperforms prehensile manipulation baselines, human teleoperation, and prior representation-based policies, achieving over 25\% improvement in success rate on unseen cluttered scenes.  
After training in simulation, our policy can be zero-shot deployed in the real world. Across 10 diverse cluttered scenes, it achieves approximately 50\% success—comparable to human teleoperation—while demonstrating higher efficiency (mean execution time 42.6s vs. 55.9s). These results highlight the robustness and generalization of our method.  
Finally, we integrate the learned policy with a high-level planner and a grasping module to perform practical grocery retrieval tasks. This demonstrates the applicability and potential real-world impact of our approach.

In summary, our contributions are threefold:
\begin{itemize}
    \item We study the problem of \textbf{non-prehensile object rearrangement in cluttered scenes}, where effective manipulation requires extrinsic dexterity to selectively leverage and avoid environmental contacts.
    \item We introduce \textbf{Dynamics-Aware Policy Learning (DAPL)}, a framework that equips policies with a learned representation of contact-induced scene dynamics via a physical world model and curriculum learning, enabling extrinsic dexterity to emerge without hand-designed primitives or complex reward engineering.
    \item We extensively evaluate our approach in both simulation and the real-world, including a new benchmark \textbf{Clutter6D}, zero-shot sim-to-real transfer across diverse cluttered scenes, and deployment in a practical grocery retrieval task.
\end{itemize}

\section{Related Work}
\begin{figure*}[htbp]
  \centering
  \includegraphics[width=0.9\linewidth]{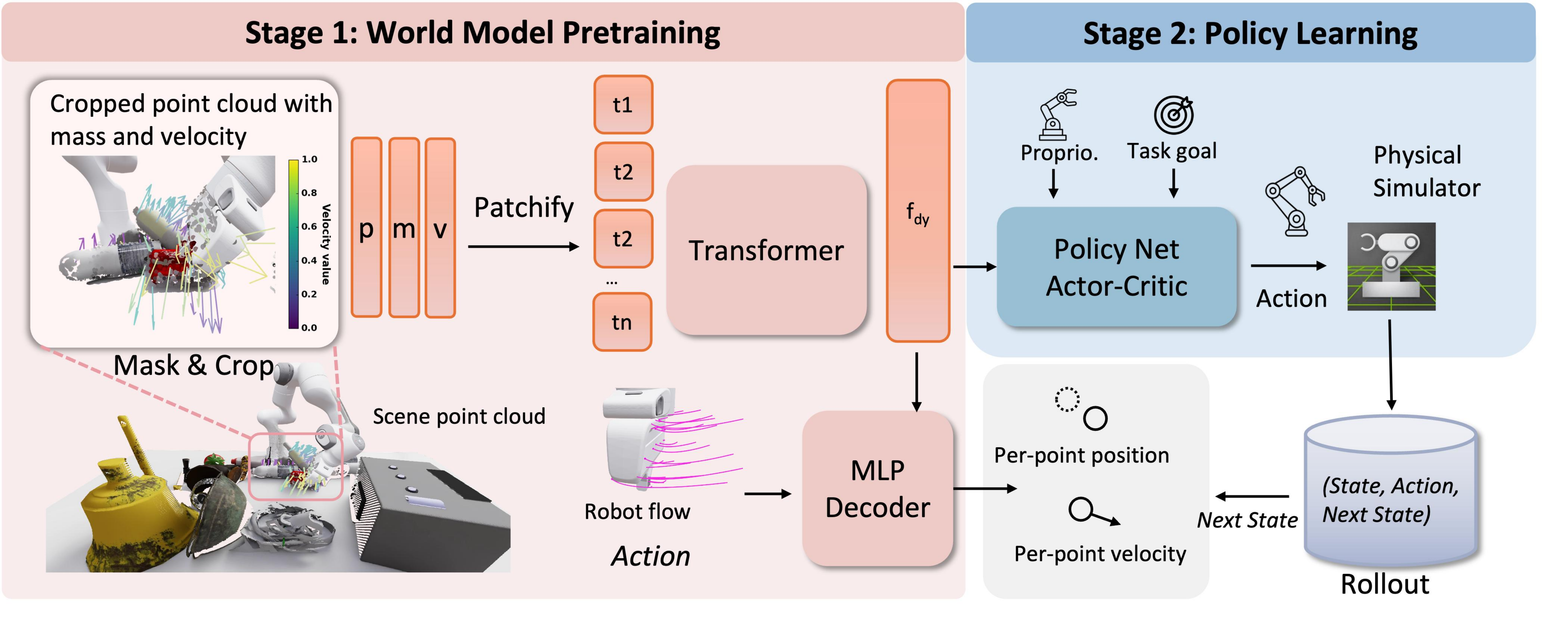}
  \caption{Overview of the proposed two-stage learning framework. Stage 1 (World Model Pretraining): The model takes a cropped point cloud augmented with physical attributes (mass and velocity) as input. A Transformer-based architecture encodes these inputs into dynamics features ($f_{dy}$), which are used by an MLP decoder to predict future per-point positions and velocities conditioned on robot actions. Stage 2 (Policy Learning): The pre-trained dynamics representations ($f_dy$) are fed into an Actor-Critic policy network alongside proprioceptive data and task goals to facilitate efficient policy learning within a physical simulator.}
  \label{fig:architecture}
\end{figure*}
\subsection{Extrinsic Dexterity}
Extrinsic dexterity\cite{dafle2014extrinsic} refers to manipulation skills that augment a robot’s intrinsic hand capabilities by exploiting external resources such as environmental contacts, gravity, and dynamic arm motions.\cite{dafle2017sampling, hou2020manipulation, hou2018fast} Prior work\cite{chavan2018hand, cheng2021contact, cheng2022contact, yuan2019end, kim2023pre} has demonstrated that complex manipulation behaviors can be achieved with simple grippers, including in-hand reorientation, prehensile pushing, and shared grasping. However, these successes largely rely on hand-crafted trajectories, task-specific motion primitives, or explicit planning over contact modes, which require precise object pose estimation and extensive manual design. As a result, such methods do not scale well beyond narrowly defined tasks or object sets. 
Learning-based methods~\cite{wu2024one,kobashi2025leveraging,yang2024learning} improve generalization, but they remain constrained to simplified scenes and are still brittle in cluttered, contact-rich environments.
Dengler et al.~\cite{dengler2024learning} study learning-based pushing in clutter; however, their method focuses on 2D collision avoidance and does not address the challenges of full 6D object rearrangement with complex, contact-rich dynamics.

\subsection{Representation Learning for Robotic Manipulation}
Prior work accelerates reinforcement learning via representation pre-training with high-dimensional observations, using pretext tasks such as geometric completion, pose prediction, or contrastive learning\cite{cho2024corn, bao2023dexart, brahmbhatt2019contactgrasp, liang2019pointnetgpd, mo2021where2act}.
Recent self-supervised\cite{zhang2025concerto, pprl, hou20254d, wu2025sonata} approaches based on patch-wise transformers learn strong geometric features for point clouds, but primarily model static shape and lack sensitivity to contact and interaction dynamics.
To better support manipulation, CORN\cite{cho2024corn} introduces contact-centric object representations, later extended by UniCORN\cite{cho2025hierarchical} to scene-level point clouds.
However, these geometry-driven representations remain brittle in dense clutter, where effective manipulation requires selectively avoiding lightweight objects while exploiting heavier ones as external supports for extrinsic dexterity.
Recent dynamics-aware pretraining methods, such as \cite{luo2024pre}, learn temporal representations from videos in pixel space, but do not model object-level physical properties or interaction forces.
As a result, they are insufficient for contact-rich robotic manipulation that hinges on physically grounded scene dynamics.

\section{Method}
Our goal is to learn a non-prehensile manipulation policy that can selectively exploit or avoid environmental contacts to achieve object rearrangement in cluttered scenes.
As illustrated in Fig.~\ref{fig:architecture}, our framework decouples the dynamics representation learning stage and RL policy learning stage.
Specifically, our framework iteratively trains:
(1) a physical world model using policy rollout interaction data to capture contact-induced object--scene dynamics; and
(2) an RL policy that explores in a physical simulator while being conditioned on the learned dynamics representation.

\subsection{Dynamics Representation Learning}
To explicitly encode this physical prior, we learn a \emph{physical world model} that predicts future object-centric dynamics conditioned on the current state and action.
The world model serves as a physics-informed representation learner whose latent features are subsequently reused for downstream policy learning.

\noindent\textbf{Physical Scene Representation.}
At each timestep $t$, the state consists of three components:
(i) the target object point cloud $\mathcal{P}^{\text{obj}}$,
(ii) the surrounding scene point cloud $\mathcal{P}^{\text{scene}}$, and
(iii) the robot end-effector point cloud.
The surrounding scene point cloud is spatially cropped within a local region centered at the target object, focusing the representation on contact-relevant geometry while reducing background redundancy.
Each 3D point is augmented with physical attributes, resulting in a per-point feature:
\begin{equation}
\mathbf{x}_i = (\mathbf{p}_i,\; m_i,\; \mathbf{v}_i),
\end{equation}
where $\mathbf{p}_i \in \mathbb{R}^3$ denotes the point position, $m_i$ the point mass, and $\mathbf{v}_i$ the point velocity.

\noindent\textbf{Architecture.}
As shown in Fig.~\ref{fig:architecture}, the world model consists of a dynamics encoder and decoder, built upon a patch-based transformer backbone.
Following prior work on point cloud transformers~\cite{cho2024corn,cho2025hierarchical,pprl}, we first partition the point cloud into local patches.
Patch centers are selected using Farthest Point Sampling (FPS), and each patch is formed by gathering the $k$-nearest neighbors (kNN) of the center point.
Each patch is normalized by subtracting its center coordinate.
A lightweight PointNet-style encoder embeds each normalized patch into a fixed-dimensional token.
To restore global spatial information, we add sinusoidal positional embeddings based on the patch center coordinates.
The resulting patch tokens are processed by a ViT, which models multi-object coupling effects.
On top of the transformer features, a small MLP decoder predicts future point-wise dynamics, including positions and velocities.

\noindent\textbf{World Model Training Objective.}
We train the world model with dense point-level supervision.
Given predicted position $\hat{\mathbf{p}}_i^{t+1}$ and velocity $\hat{\mathbf{v}}_i^{t+1}$, we minimize
\begin{equation}
\mathcal{L}_{\text{dyn}}
=
\sum_i
\left\|
\hat{\mathbf{p}}_i^{t+1} - \mathbf{p}_i^{t+1}
\right\|_2^2
+
\lambda
\left\|
\hat{\mathbf{v}}_i^{t+1} - \mathbf{v}_i^{t+1}
\right\|_2^2 .
\end{equation}
In cluttered scenes, most points exhibit near-zero velocities, leading to highly sparse supervision in the velocity field. As a result, optimizing only point-wise velocity losses can cause the model to collapse to trivial solutions that predict uniformly small velocities and fail to capture contact-induced motion. To address this, we treat the scene as a point distribution rather than independent points and introduce a variance-aware regularization over point-wise velocities. Specifically, we match the standard deviations of the predicted and ground-truth velocity fields:
\begin{equation} \mathcal{L}_{\text{var}} = \left\| \mathrm{Std}\!\left(\{\hat{\mathbf{v}}_i^{t+1}\}_i\right) - \mathrm{Std}\!\left(\{\mathbf{v}_i^{t+1}\}_i\right) \right\|_2 , \end{equation}
This regularization preserves the overall magnitude and spatial variability of motion in dynamic regions, preventing degenerate near-zero velocity collapse.

The overall world model training objective is defined as
\begin{equation}
\mathcal{L}
=
\mathcal{L}_{\text{dyn}}
+
\alpha \mathcal{L}_{\text{var}},
\end{equation}
where $\alpha$ controls the strength of variance-aware regularization. 

\subsection{Dexterous Policy Learning via RL}

\noindent\textbf{Observation and Action Spaces.}
At each timestep, the policy observes three components: a dynamics-aware scene representation, the robot proprioceptive state, and a task goal. The physical scene is processed by the dynamics encoder to produce a set of embeddings that capture contact-induced object dynamics. The robot proprioceptive observation consists of joint positions, joint velocities, and end-effector poses, while the task goal is represented as the relative pose between the target object and the desired goal configuration. The policy outputs continuous joint-space control commands, which are executed using an impedance controller.

\noindent\textbf{Reward Design.} 
Our reward design follows standard practice in non-prehensile manipulation (e.g., HAMNet~\cite{cho2025hierarchical}) and uses only lightweight shaping rather than heavy task-specific reward engineering.
The reward consists of a sparse task success term and two lightweight shaping terms that encourage physical contact and goal-directed object motion.
The contact term $r_{\text{contact}}$ is defined based on the minimum distance $d_{\text{oe}}$ between the target object and the end-effector, which encourages interaction:
\begin{equation}
r_{\text{contact}} = 1 - \tanh(d_{\text{oe}}).
\end{equation}
The goal-reaching term $r_{\text{goal}}$ measures the distance $d_{\text{og}}$ between the object and the desired goal pose, and is activated once the end-effector distance $d_{\text{oe}}$ falls below a threshold $\tau_d$:
\begin{equation}
r_{\text{goal}} = \mathbb{I}(d_{\text{oe}} < \tau_d)\left(1 - \tanh(d_{\text{og}})\right).
\end{equation}
A sparse success reward is issued when the target object reaches the desired pose. To penalize disturbance to the scene, the reward is discounted by the displacement of non-target objects $D_{\text{disp}}$ measured by Chamfer distance:
\begin{equation}
r_{\text{success}} = \mathbb{I}_{\text{success}} \left(1 - \beta D_{\text{disp}}\right)
\end{equation}
All coefficients and thresholds use standard values and are provided in the Appendix.

\subsection{Curriculum Learning with Policy Interaction}
Rather than relying on a fixed offline dataset, we adopt an iterative curriculum that alternates between policy learning and world model refinement.
We initialize the process by training an RL policy from scratch without a pretrained dynamics representation.
In each curriculum iteration, we freeze the dynamics encoder and train the policy until both reward and success rate plateau. This typically takes about $2\times10^4$ RL steps per stage, with fewer steps required in later iterations.
Once the policy reaches basic task coverage, we roll out the policy to collect approximately $60\mathrm{k}$ interaction steps.
These trajectories are intentionally imperfect and include substantial random collisions and suboptimal behaviors, which is beneficial for exposing diverse contact dynamics.
The collected interaction data is then used to update the world model for $500\mathrm{k}$ training steps until the validation loss stabilizes, improving its ability to capture contact-induced momentum transfer under realistic policy-induced distributions.
The refined dynamics encoder is subsequently reused by the RL policy, yielding improved exploration efficiency and more stable policy learning.
We repeat this alternating procedure until both the policy performance and the learned dynamics representation converge.
This curriculum allows the world model and policy to co-evolve, progressively shifting from noisy exploratory interactions to task-relevant, physically consistent manipulation behaviors.

\subsection{Clutter6D: Object Rearrangement Environment}

To train and evaluate manipulation policy in clutter scenes, we introduce \textbf{Clutter6D}, a comprehensive 6D object rearrangement environment and benchmark. 
Clutter6D is designed to explicitly stress manipulation scenarios where extrinsic dexterity is necessary rather than optional. Increasing clutter density not only reduces free-space motion but also amplifies the coupling between object dynamics, making contact outcomes highly sensitive to physical properties. In such regimes, collision-free planning and grasp-centric strategies become unreliable, and successful manipulation requires selectively leveraging or avoiding environmental contacts.
Compared to prior clutter benchmarks that emphasize collision avoidance or planar pushing, Clutter6D focuses on full 6D object rearrangement with multi-object contact and dynamic coupling. The use of task-oriented scene graphs enables systematic control over relational constraints while preserving scene diversity, making the benchmark suitable for evaluating contact-rich manipulation policies beyond grasping.

\noindent\textbf{Environment Setup.}
Clutter6D is built on the IsaacLab simulator with PhysX as the physics backend. We curate high-quality object assets from Objaverse~\cite{deitke2023objaverse} and process them using mesh simplification and convex decomposition (PaMO\cite{oh2025pamo} and CoACD\cite{wei2022coacd}). Physical properties such as scale and mass, as well as semantic attributes including category, color, and material, are automatically generated\cite{gpt5} and verified through a human-in-the-loop process. From these, we select and normalize 10K assets to tabletop scale. Task scenarios are synthesized from object semantics and represented as task-oriented scene graphs (ToSG)\cite{gao2025genmanip}, which specify spatial relations among objects. Scenes are instantiated by placing objects in topological order according to the scene graph, discarding placements that violate relational constraints.

\noindent\textbf{Benchmark Setting.}
Clutter6D defines three difficulty tracks based on clutter density: \textit{Sparse} (4 objects), \textit{Moderate} (8 objects), and \textit{Dense} (12 objects). The sparse track contains 1,024 scenes for training, while each track includes 128 held-out scenes for evaluation. Performance is evaluated using two metrics: (i) task success rate, where success is achieved if the target object reaches the desired pose within 0.05 m and 0.1 rad, and (ii) chamfer distance of non-target object offsets, measuring unintended disturbance to the surrounding clutter. Episodes terminate upon task completion, object drop, or after 300 simulation steps.

\section{Experiments}
\begin{figure*}[htbp]
  \centering
  \includegraphics[width=0.9\linewidth]{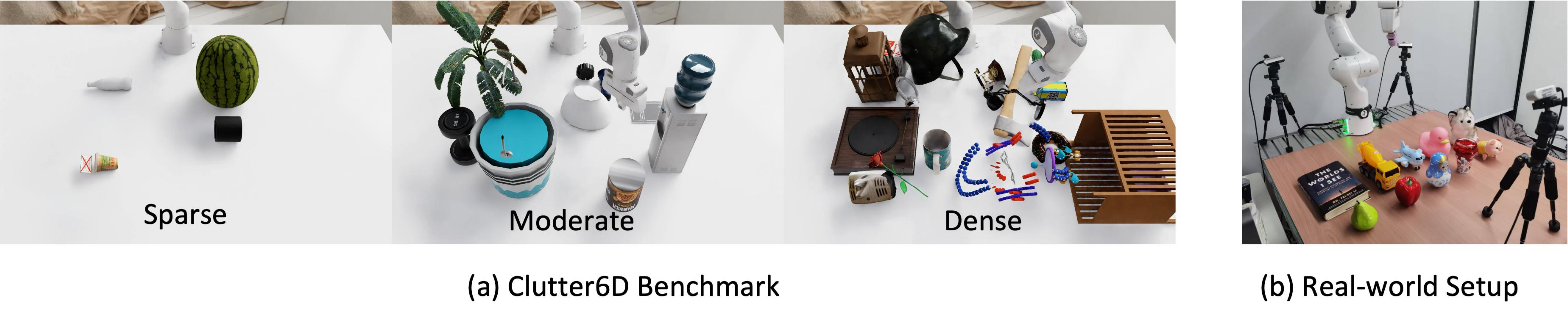}
  \vspace{-0.4cm}
  \caption{Overview of our proposed Clutter6D Benchmark and real-world setup. (a) Representative examples of the Clutter6D Benchmark with three levels of scene density: Sparse (4 objects), Moderate (8 objects), and Dense (12 objects). (b) The real-world experimental setup consisting of a Franka Research 3 robot and three Intel RealSense cameras for point cloud acquisition, along with the complete set of objects used in our real-world experiments.}
  \label{fig:reward}
\end{figure*}
In this section, we evaluate our method in both the proposed Clutter6D simulation benchmark and real-world scenarios. 

\subsection{Simulation Experiments}
\noindent\textbf{Setup and Baselines.}
We evaluate all methods on the Clutter6D benchmark across three difficulty levels: \textit{Sparse} (4 objects), \textit{Moderate} (8 objects), and \textit{Dense} (12 objects). We compare our approach against three categories of baselines:
\begin{itemize}
    \item \textbf{Prehensile Manipulation:} GraspGen + CuRobo \cite{murali2025graspgen, sundaralingam2023curobo}, a pipeline combining grasp synthesis with motion planning.
    \item \textbf{Human Teleoperation:} A baseline established by expert human operators using a Gello \cite{wu2024gello} interface.
    \item \textbf{Representation Learning+RL Policy:} 
    We compare our method against two types of learning-based baselines: (i) general point cloud encoders, including Point2Vec \cite{abouzeid2023point2vec} and Concerto \cite{zhang2025concerto}, and (ii) encoders specifically tailored for non-prehensile manipulation: CORN \cite{cho2024corn} and UniCORN\cite{cho2025hierarchical}. We also include a CORN-multi baseline which uses a CORN encoder to extract features from each individual object and fuses them into a global representation. 
\end{itemize}

All learning-based methods share comparable policy networks to ensure a fair comparison. While the baselines use only standard geometric coordinates $(x, y, z)$ as input, our approach leverages the full 7-dimensional physical features (including velocity and mass), highlighting the advantage of explicit physical reasoning and a dynamics-aware learning paradigm rather than architectural differences.

\begin{table*}[t]
    \setlength{\tabcolsep}{5pt}
    \centering
    \resizebox{1.6\columnwidth}{!}
    {
    \begin{tabular}{llcccccc}
    \toprule
        \multicolumn{1}{l}{\multirow{2}{*}{\textbf{Methods}}} & \multicolumn{1}{l}{\multirow{2}{*}{\textbf{Action Type}}} & \multicolumn{2}{c}{\textbf{Sparse}} & \multicolumn{2}{c}{\textbf{Moderate}} & \multicolumn{2}{c}{\textbf{Dense}}\\ 
        \cmidrule(lr){3-4} \cmidrule(lr){5-6} \cmidrule(lr){7-8}
        \rowcolor{gray!10}
        ~ & ~ & S.R. $\uparrow$ & M.O. $\downarrow$  & S.R. $\uparrow$& M.O. $\downarrow$& S.R. $\uparrow$& M.O. $\downarrow$\\ \midrule
        Teleoperation \cite{wu2024gello} &Mixed & 50.0 & 3.13 & 40.0 & 7.49 & 20.0 & 21.34\\
        GraspGen + CuRobo \cite{murali2025graspgen, sundaralingam2023curobo} &Prehensile & 26.6 & -- & 15.6 & -- & 3.13 & -- \\ \midrule
        Point2Vec  \cite{abouzeid2023point2vec} &Non-prehensile & 6.89 & 5.09 & 1.95 & 3.36 & 0.78 &5.35   \\ 
        Concerto  \cite{zhang2025concerto} &Non-prehensile & 3.13 & 1.65 & 1.56 & 2.90 & 0.39 & 7.56  \\ 
        CORN \cite{cho2024corn} &Non-prehensile & 46.63 & 3.15 & 45.83 & 5.51 & 22.22 & 17.43  \\ 
        CORN-multi  &Non-prehensile  & 35.93  & 2.73 & 15.38 & 3.92 & 11.83 & 12.06 \\ 
        UniCORN\cite{cho2025hierarchical}  &Non-prehensile  & 20.61 & 1.71 & 11.67 & 4.13 & 5.81 & 9.79 \\
        \rowcolor{lightblue}
        Ours &Non-prehensile & \textbf{71.88} & \textbf{2.59} &\textbf{51.04} &\textbf{2.7} & \textbf{44.56} & \textbf{12.65} \\ 
        \bottomrule
    \end{tabular}}
    \vspace{-2mm}
    \caption{Quantitative results measured by success rate in the simulation benchmark. Note that Mean Offset (M.O.) for GraspGen + CuRobo is omitted because it relies on the CuRobo motion planner, which strictly prioritizes collision-free trajectories. Our method significantly outperforms all baselines across all clutter densities, maintaining high success rates even in dense environments where baselines experience a sharp performance decline.}
    \label{tab:sim_results}
\end{table*}
\begin{figure}[htbp]
  \centering
  \includegraphics[width=1.0\linewidth]{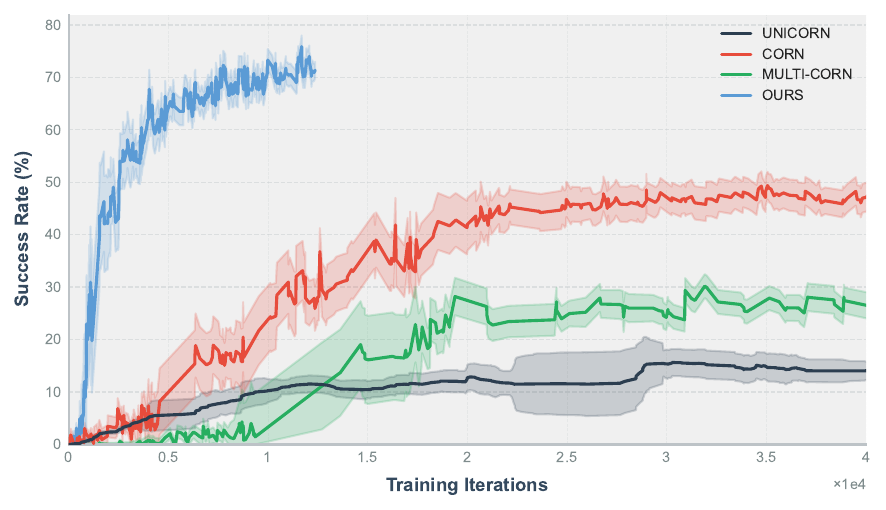}
  \vspace{-0.8cm}
  \caption{Training efficiency and convergence comparison. Our method (blue) achieves significantly higher sample efficiency compared to geometry-based baselines, reaching about 70\% success rate within the initial $10^4$ iterations. This demonstrates that our dynamics-aware representation provides a robust physical prior, facilitating the rapid acquisition of complex interaction strategies without the need for exhaustive sampling of contact physics.}
  \label{fig:training_curve}
\end{figure}

\begin{figure}[htbp]
  \centering
  \includegraphics[width=1.0\linewidth]{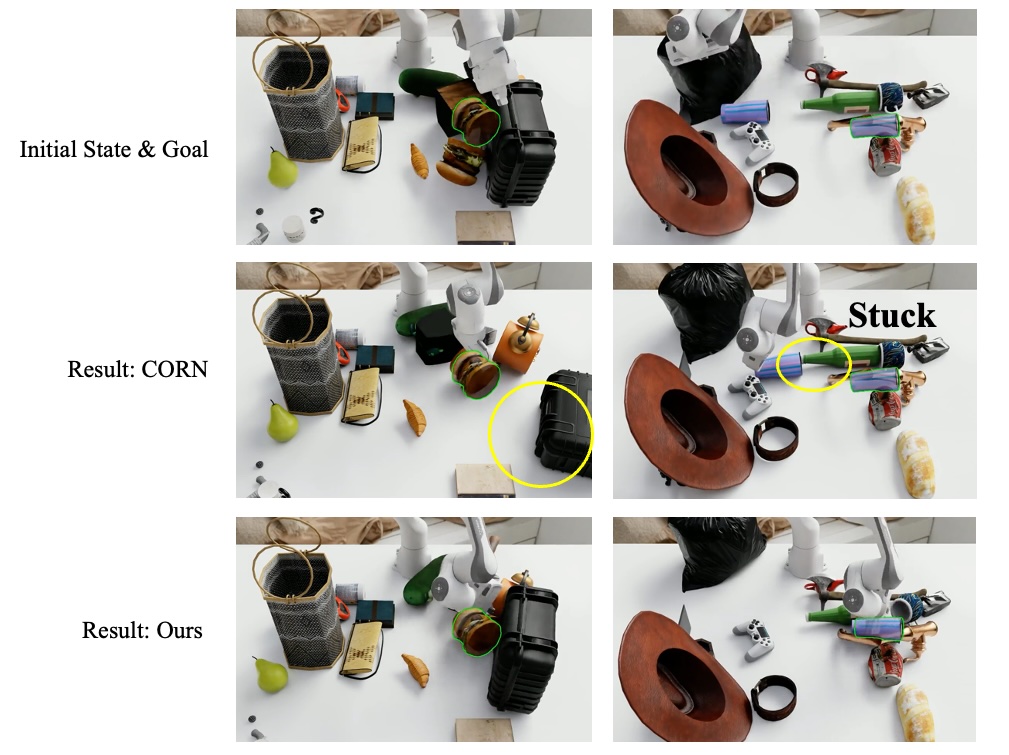}
  \caption{Compared to static geometric representation baseline CORN, which either triggers excessive contact chains leading to significant environment disturbance (middle left) or becomes "stuck" due to a lack of dynamics reasoning (middle right), our policy (bottom row) exhibits selective and intentional interactions, successfully reorients the target while preserving scene stability.}
  \label{fig:comparion with baseline}
\end{figure}
\begin{figure*}[ht]
  \centering
  \includegraphics[width=1.0\linewidth]{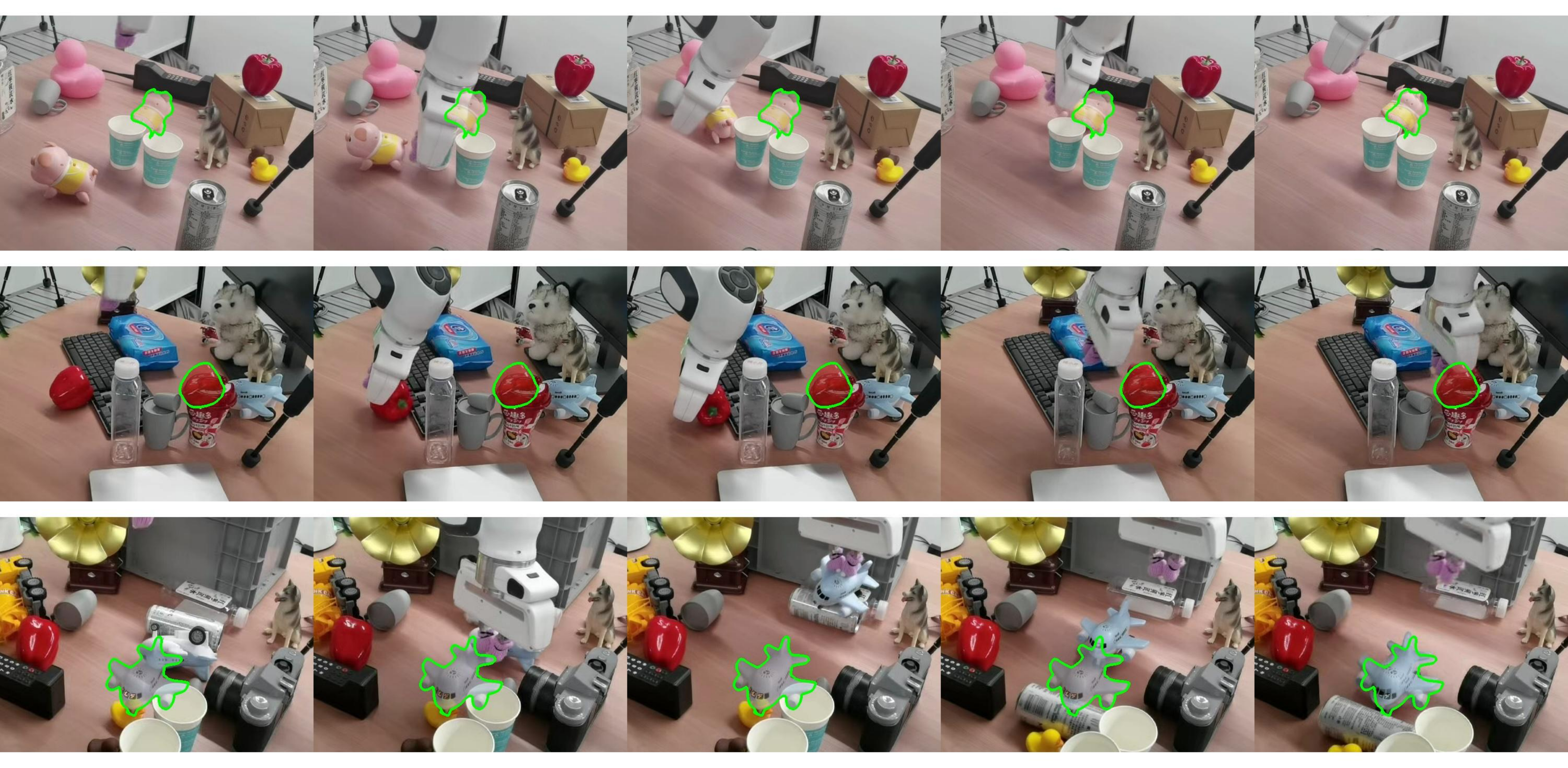}
  \vspace{-0.5cm}
  \caption{Qualitative Results in the real world. The goal pose is shown transparently.}
  \label{fig:realworld_frames}
\end{figure*}

\noindent\textbf{Main Results.}
Table~\ref{tab:sim_results} summarizes quantitative results across varying clutter densities, while Fig.\ref{fig:realworld_frames} shows representative qualitative rollouts. Our method consistently outperforms all baselines, with the performance gap becoming particularly pronounced in the \textit{Dense} setting. While geometry-based methods such as CORN achieve reasonable success in sparse scenes (46.63\%), their performance degrades sharply under heavy clutter (22.22\%). In contrast, our approach maintains a robust success rate of \textbf{44.56\%} in dense environments, effectively doubling the strongest baseline.

This advantage is further reflected in the Mean Offset (M.O., measured in cm) metric, which measures unintended disturbance to non-target objects. Our method achieves a favorable balance between task success and environment preservation. Specifically, while the CORN baseline manages a certain level of success, it incurs significantly higher disturbance (17.43 vs. our 12.65) due to its lack of dynamics reasoning. Conversely, lower offsets reported by other methods are largely artifacts of execution failure, whereas our policy maintains the highest success rate through precise, intentional interactions. Qualitatively, as shown in Fig.\ref{fig:comparion with baseline}, CORN relies on static geometric reasoning and often initiates excessive or poorly directed contacts, triggering unintended contact chains that cause the robot to become stuck in clutter. In contrast, our policy exhibits more selective and intentional interactions, exploiting environmental contacts when beneficial while actively avoiding unnecessary collisions. Together, these results support our core hypothesis that static geometric representations are insufficient in dense clutter, where momentum transfer and multi-object contact dynamics critically determine task success.

\noindent\textbf{Training Efficiency and Convergence.}
Beyond final performance, Fig.~\ref{fig:training_curve} illustrates the learning dynamics of different methods. Our approach demonstrates significantly higher sample efficiency, reaching approximately 70\% success within the first few thousand training iterations. In contrast, baselines relying on static geometric representations (e.g., CORN and UniCORN) converge substantially slower, while other representations fail to learn meaningful policies. These results indicate that the dynamics-aware representation provides a strong physical prior, allowing the policy to learn effective interaction strategies early without expending excessive samples on discovering basic contact physics.


\setlength{\tabcolsep}{3pt}
\begin{table}[t]
    \centering
    \resizebox{\columnwidth}{!}
    {
    \begin{tabular}{lcccccc}
    \toprule
    \rowcolor{gray!10}
        Pretrain Task & Granularity  & Velocity & Phys. & P.E. $\downarrow$ & S.R. $\uparrow$& M.O. $\downarrow$ \\
    \midrule
        Recons. & Point-level & \xmark & \xmark &- &11.75 &1.31\\ 
        Recons. & Point-level & \cmark & \cmark &- &29.63 &2.63\\
        World Model & Object-level           & \xmark & \xmark & 3.1 & 14.13 & 3.27 \\ 
        World Model & Object-level   & \cmark & \cmark & 3.2 & 16.88 & 3.84 \\ 
        World Model & Point-level            & \xmark & \xmark & 4.1 & 42.00 & 4.91 \\  
        World Model & Point-level           & \cmark & \xmark & 5.1 & 58.25 & 4.86 \\ 
        \rowcolor{lightblue}
        World Model & Point-level  & \cmark & \cmark & \textbf{4.6} & \textbf{71.88} & \textbf{2.59} \\ 
    \bottomrule
    \end{tabular}}
    \caption{Ablation Study on Pre-training Objectives and Input Modalities (Sparse Track). P.E.: Chamfer Distance prediction error (cm).  Results highlight the indispensability of velocity and physical features, while further establishing that point-level dynamics modeling is a more effective pretext task than either object-level relative pose prediction or simple reconstruction.
}
\label{tab:ablation}
\end{table}

\begin{figure}[htbp]
  \centering
  \includegraphics[width=1.0\linewidth]{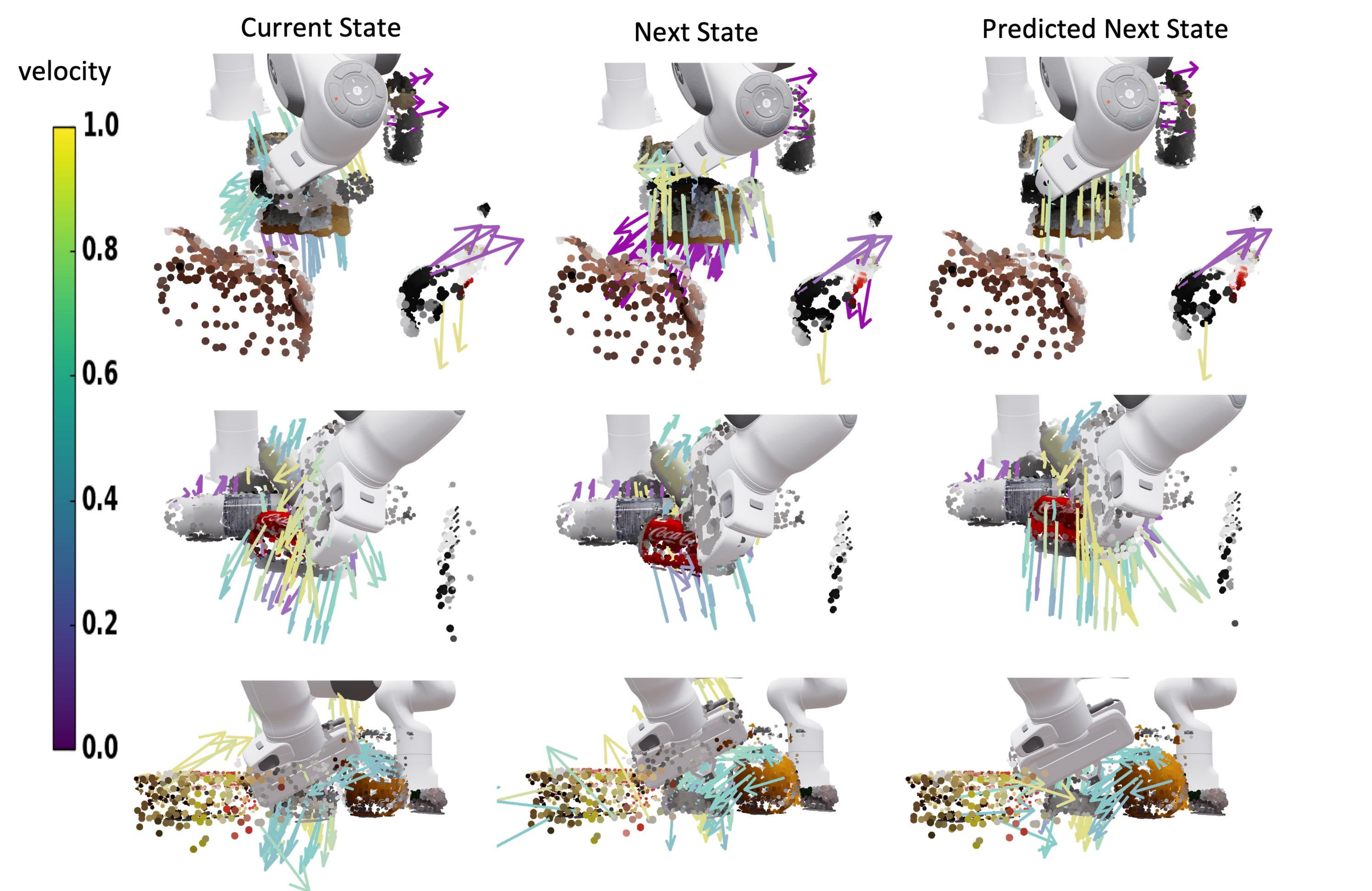}
  \caption{Visualization of our world model predictions. }
  \label{fig:visualization}
\end{figure}
\noindent\textbf{Analysis of Dynamics Representation.}
We conduct ablation studies to assess the contribution of the proposed dynamics representation (Table~\ref{tab:ablation}). Removing physical attributes such as velocity and mass leads to a substantial performance drop, highlighting the importance of reasoning about the \textit{potential for motion} in cluttered manipulation. Additionally, replacing point-level dense prediction with object-level 6 DoF pose prediction leads to inferior performance, suggesting that pose supervision is too coarse to learn the intricate contact physics and local deformations essential for precise tracking. Moreover, replacing the dynamics prediction objective with simple autoencoder reconstruction significantly degrades performance, indicating that the effectiveness comes from dynamics learning other than static shape reconstruction. We further visualize world model predictions and observe close agreement with ground-truth object motion in Fig.\ref{fig:visualization}, confirming the quality of learned dynamics. 

\begin{figure}[htbp]
  \centering
  \includegraphics[width=1.0\linewidth]{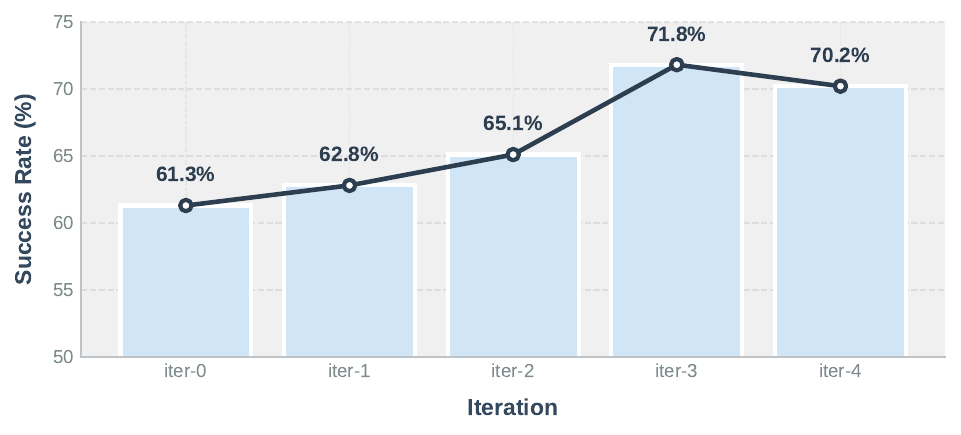}
  \caption{Success rate over curriculum iterations. Performance steadily improves from 61.3\% to 71.8\%, demonstrating that iterative model refinement provides more accurate dynamics conditioning for learning effective manipulation strategies.}
  \label{fig:curriculum}
\end{figure}

\noindent\textbf{Effectiveness of Curriculum Learning.}
Fig.~\ref{fig:curriculum} shows the evolution of policy performance across curriculum iterations. Starting from 61.3\%, the success rate steadily improves to \textbf{71.8\%} after three iterations. Early policies generate diverse and imperfect interaction trajectories, which are used to refine the world model. The improved dynamics representation then provides more accurate conditioning for subsequent policy learning, leading to progressively more effective extrinsic dexterity.

\noindent\textbf{Behavior Analysis for Extrinsic Dexterity.}
To examine whether the learned policy exhibits extrinsic dexterity beyond aggregate success metrics, we manually analyze 50 sampled simulation trajectories for each method. We measure unnecessary contacts, disturbance to non-target objects, and the occurrence of Avoid, Traverse, and Leverage behaviors. Table~\ref{tab:behavior_analysis} summarizes the results.
\begin{table}[t]
    \centering
    \scriptsize
    \setlength{\tabcolsep}{2.4pt}
    \renewcommand{\arraystretch}{1.08}
    \resizebox{\columnwidth}{!}{%
    \begin{tabular}{@{}lccccccccc@{}}
    \toprule
    \rowcolor{gray!10}
    \multirow{2}{*}{Method} &
    \multirow{2}{*}{\shortstack{Unnec.\\Contacts}} &
    \multicolumn{4}{c}{Object Disturbance} &
    \multicolumn{4}{c}{Dexterous Behaviors (\%)} \\
    \cmidrule(lr){3-6} \cmidrule(lr){7-10}
    \rowcolor{gray!10}
    & & \shortstack{Small\\Trans.} & \shortstack{Small\\Rot.} & \shortstack{Large\\Trans.} & \shortstack{Large\\Rot.} & Avoid & Traverse & Leverage & Simple \\
    \midrule
    CORN & 3.6 & 2.8 & 2.6 & 4.4 & 4.8 & 22 & 14 & 6 & 62 \\
    \rowcolor{lightblue}
    Ours & \textbf{2.4} & 3.2 & 3.0 & \textbf{1.8} & \textbf{1.2} & \textbf{52} & \textbf{38} & \textbf{42} & 28 \\
    \bottomrule
    \end{tabular}%
    }
    \caption{Behavior analysis over 50 sampled trajectories per method. Unnec. Contacts and Object Disturbance metrics are reported as average counts per trajectory. Small/Large Trans. count non-target objects with translations below/above 5 cm, while Small/Large Rot. count non-target objects with rotations below/above $45^\circ$. Avoid, Traverse, and Leverage correspond to the dexterous behavior types illustrated in Fig.~\ref{first_figure}. Multiple behavior types may occur in one trajectory; Simple denotes trajectories in which none of the three is observed.}
    \label{tab:behavior_analysis}
\end{table}

Compared with CORN, DAPL produces fewer unnecessary contacts and substantially fewer large translations and rotations of non-target objects. It also exhibits Avoid, Traverse, and Leverage behaviors more frequently, suggesting that the learned policy develops purposeful contact strategies rather than relying on indiscriminate pushing.

\noindent\textbf{Sensitivity to Noisy Inputs.}
We further evaluate policy sensitivity to noisy physical and goal inputs through controlled perturbations in simulation. As shown in Table~\ref{tab:noise_robustness}, mass and velocity perturbations maintain similar success rates over a broad noise range, although larger noise gradually increases scene disturbance. In contrast, goal-pose noise causes a much sharper performance drop because it directly changes the task specification.
\begin{table}[t]
    \centering
    \scriptsize
    \setlength{\tabcolsep}{2.6pt}
    \renewcommand{\arraystretch}{1.08}
    \resizebox{0.92\columnwidth}{!}{%
    \begin{tabular}{@{}lcccccccc@{}}
    \toprule
    \rowcolor{gray!10}
    \multirow{2}{*}{Noise Source} &
    \multicolumn{2}{c}{0.5\%} &
    \multicolumn{2}{c}{5\%} &
    \multicolumn{2}{c}{25\%} &
    \multicolumn{2}{c}{50\%} \\
    \cmidrule(lr){2-3} \cmidrule(lr){4-5} \cmidrule(lr){6-7} \cmidrule(lr){8-9}
    \rowcolor{gray!10}
    & S.R. & M.O. & S.R. & M.O. & S.R. & M.O. & S.R. & M.O. \\
    \midrule
    Mass & 66.9 & 3.1 & 67.3 & 3.4 & 66.0 & 4.1 & 62.4 & 5.9 \\
    Velocity & 64.0 & 3.6 & 64.0 & 3.2 & 61.1 & 3.9 & 58.5 & 4.4 \\
    Goal Pose & 66.3 & 3.1 & 65.0 & 3.3 & 48.1 & 3.5 & 15.1 & 4.0 \\
    All Inputs & 65.7 & 3.7 & 66.6 & 4.9 & 45.1 & 6.0 & 5.9 & 8.4 \\
    \bottomrule
    \end{tabular}%
    }
    \caption{Sensitivity to noisy physical and goal inputs under controlled simulation perturbations. Noise levels are reported as percentages of the corresponding input mean. S.R. denotes success rate and M.O. denotes mean offset (cm).}
    \label{tab:noise_robustness}
\end{table}


\noindent\textbf{Analysis of Adaptive Behavior.}
\begin{figure}
    \centering
    \includegraphics[width=1\linewidth]{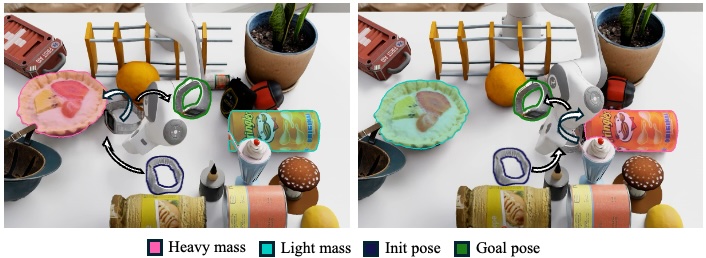}
    \caption{Adaptive behavior guided by physical priors. In identical scenes, our policy autonomously adapts its interaction trajectory based on whether the pie or the Pringles can (pink/cyan mask) is assigned a heavy mass, transitioning between using objects as stable anchors versus avoiding them to prevent destabilization}
    \label{fig:adaptive_behavior}
\end{figure}
Beyond quantitative performance gains, our policy exhibits adaptive interaction strategies that reflect a deep understanding of contact-induced dynamics. We demonstrate this through a controlled experiment where we maintain identical initial and goal configurations while modifying the objects' physical properties. As shown in Fig.\ref{fig:adaptive_behavior}, when the pie is heavy and the Pringles can is light, the policy proactively exploits the pie as a stable anchor. It establishes deliberate contact with the pie to generate the necessary torque for object reorientation. Conversely, if the mass assignments are swapped, the policy autonomously adapts its trajectory to avoid the lightweight pie. It prevents unintended scene destabilization by utilizing the now-heavy Pringles can as the functional support instead. These results suggest that the policy reasons about how interactions unfold purely through a dynamics-aware representation rather than relying on geometric heuristics.

\subsection{Real-world Experiments}
\noindent\textbf{System Setups.}
We deploy our policy on a Franka Research 3 robot equipped with three Intel RealSense cameras. Object segmentation is initialized using SAM2\cite{sam} on the first frame and tracked online with XMem\cite{xmem}. Object poses are estimated using FoundationPose\cite{foundationpose}.
For sim-to-real transfer, we perform system identification of key physical parameters. Object mass is estimated using a vision–language model\cite{gpt5} that provides coarse physical priors, while velocities are obtained by temporal differentiation followed by EKF-based filtering~\cite{EKF}. To further bridge the sim-to-real gap, we employ a student-teacher distillation framework under perturbation. By training the student to recover the teacher's latent dynamics from observations injected with Gaussian noise, we force the policy to infer effective dynamics despite imperfect perception. Although these estimates are not precise, they provide sufficient priors for the policy to select appropriate interaction strategies. We compare our method against FACTR \cite{liu2025factr}, a teleoperation baseline that enhances operator precision in contact-rich tasks through multimodal sensory integration. We evaluate performance on 10 diverse real-world cluttered scenes. A trial is considered successful if the target object reaches the desired pose with errors below 0.05 m and 0.1 rad within 90 seconds. We report success rate and mean execution time.

\begin{table}[t]
    \setlength{\tabcolsep}{2.5pt}
    \centering
    \resizebox{\columnwidth}{!}{%
    \begin{tabular}{@{}lcccccccc@{}}
    \toprule
    \textbf{Methods}
        & \multicolumn{2}{c}{S1}
        & \multicolumn{2}{c}{S2}
        & \multicolumn{2}{c}{S3}
        & \multicolumn{2}{c}{S4} \\
    \cmidrule(lr){2-3} \cmidrule(lr){4-5} \cmidrule(lr){6-7} \cmidrule(lr){8-9}
    \rowcolor{gray!10}
        & S.R. $\uparrow$ & M.T. $\downarrow$
        & S.R. $\uparrow$ & M.T. $\downarrow$
        & S.R. $\uparrow$ & M.T. $\downarrow$
        & S.R. $\uparrow$ & M.T. $\downarrow$ \\
    \midrule
    Teleop.~\cite{liu2025factr}
        & 3/5 & 77 & 4/5 & 59 & 3/5 & 39 & 1/5 & 81 \\
    \rowcolor{lightblue}
    Ours
        & \textbf{2/5} & \textbf{51}
        & \textbf{1/5} & \textbf{26}
        & \textbf{3/5} & \textbf{33}
        & \textbf{1/5} & 73 \\
    \midrule[\heavyrulewidth]
    \textbf{Methods}
        & \multicolumn{2}{c}{S5}
        & \multicolumn{2}{c}{S6}
        & \multicolumn{2}{c}{S7}
        & \multicolumn{2}{c}{S8} \\
    \cmidrule(lr){2-3} \cmidrule(lr){4-5} \cmidrule(lr){6-7} \cmidrule(lr){8-9}
    \rowcolor{gray!10}
        & S.R. $\uparrow$ & M.T. $\downarrow$
        & S.R. $\uparrow$ & M.T. $\downarrow$
        & S.R. $\uparrow$ & M.T. $\downarrow$
        & S.R. $\uparrow$ & M.T. $\downarrow$ \\
    \midrule
    Teleop.~\cite{liu2025factr}
        & 3/5 & 28 & 2/5 & 39 & 4/5 & 20 & 0/5 & 90 \\
    \rowcolor{lightblue}
    Ours
        & 4/5 & 37 & 3/5 & 59 & 2/5 & 24 & 2/5 & 28  \\
    \midrule[\heavyrulewidth]
    \textbf{Methods}
        & \multicolumn{2}{c}{S9}
        & \multicolumn{2}{c}{S10}
        & \multicolumn{2}{c}{Avg.}
        & \multicolumn{2}{c}{} \\
    \cmidrule(lr){2-3} \cmidrule(lr){4-5} \cmidrule(lr){6-7} 
    \rowcolor{gray!10}
        & S.R. $\uparrow$ & M.T. $\downarrow$
        & S.R. $\uparrow$ & M.T. $\downarrow$
        & S.R. $\uparrow$ & M.T. $\downarrow$
        & & \\
    \midrule
    Teleop.~\cite{liu2025factr}
        & 5/5 & 62 & 1/5 & 64 & 52\% & 55.9 & & \\
    \rowcolor{lightblue}
    Ours
        & 5/5 & 45 & 1/5 & 50 & 48\% & 42.6 & & \\
    \bottomrule
    \end{tabular}%
    }
    \caption{
    Quantitative results across 10 scenes.
    Mean (last column) denotes average performance across all scenes.
    SR $\uparrow$ indicates success rate and MT $\downarrow$ indicates mean execution time.
    }
    \label{tab:realworld}
\end{table}
\begin{figure*}[ht]
  \centering
  \includegraphics[width=0.9\linewidth]{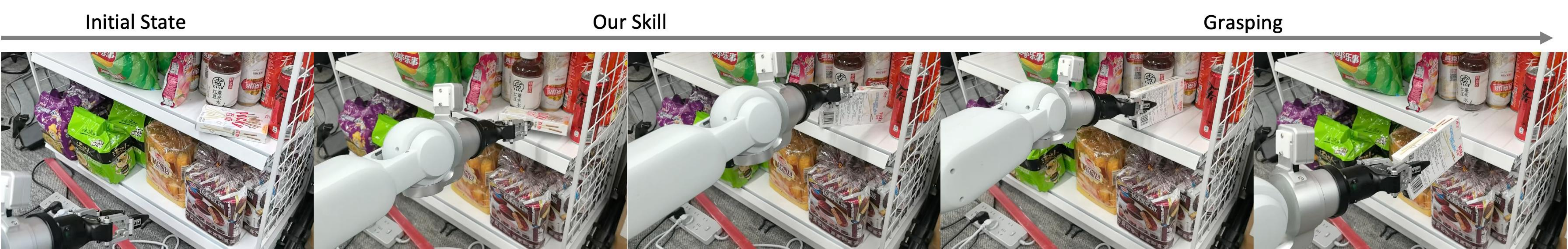}
  \caption{Humanoid grocery retrieval using extrinsic dexterity. Our policy enables the Galbot G1 to manipulate objects in cluttered shelves by sliding and reorienting them into graspable configurations}
  \label{fig:grocery}
\end{figure*}
\noindent\textbf{Results.}
As shown in Table~\ref{tab:realworld}, our method achieves an average success rate of 48\% across 10 scenes, approaching human teleoperation performance (52\%). Notably, our approach achieves comparable success with more consistent execution times, demonstrating reliable real-world performance despite sensing noise and modeling inaccuracies.
Although the dynamics representation relies on estimated physical attributes such as mass and velocity, precise estimation of these quantities is not required for effective deployment. In practice, coarse mass priors and noisy velocity estimates provide sufficient cues for the policy to differentiate stable supports from objects likely to be displaced. This observation is consistent with the sensitivity analysis in Table~\ref{tab:noise_robustness}, where mass and velocity perturbations have limited impact on success rate compared with goal-pose noise.

\subsection{Applications}
To demonstrate practical utility in unstructured environments, we deploy our policy on a humanoid robot (Galbot G1) for a grocery retrieval task.
The manipulation policy is trained entirely in a digital twin of the target shelf environment using Galbot, without access to real-world interaction data.
We integrate a VLM-based planner (SoFar\cite{qi2025sofar}) to translate natural language commands (e.g., ``\textit{Retrieve the cracker box}'') into 6D target poses. As shown in Fig.~\ref{fig:grocery}, direct grasping is often infeasible in shelf scenarios due to limited reachability or objects exceeding the gripper's width. In this pipeline, our learned policy acts as a critical \textit{pre-grasping} skill: it leverages extrinsic dexterity to slide and reorient the target object out of the clutter into a grasp-friendly configuration, enabling the downstream grasping module\cite{graspvla} to successfully complete the task. 
\section{Conclusion, Limitations and Future Direction} 
\label{sec:conclusion}

We studied non-prehensile object rearrangement in cluttered scenes, where effective manipulation requires extrinsic dexterity to selectively leverage and avoid environmental contacts. We proposed \textbf{Dynamics-Aware Policy Learning (DAPL)}, which equips reinforcement learning policies with a learned representation of contact-induced scene dynamics via a physical world model and curriculum learning. Furthermore, we introduced Clutter6D, a new benchmark tailored for evaluating 6-DoF non-prehensile manipulation across varying clutter densities, where DAPL achieves state-of-the-art performance. By reasoning about interaction outcomes beyond static geometry, our method enables robust extrinsic dexterity; importantly, in real-world experiments, DAPL achieves success rates comparable to human teleoperation while noticeably reducing average execution time.

Our approach has several limitations. The dynamics representation relies on approximate physical attributes (e.g., object mass and velocity), which may be noisy or unavailable in real-world settings. In addition, we focus on tabletop rearrangement with rigid objects and do not explicitly address articulated or deformable objects, nor long-horizon multi-stage tasks. 
Future work includes improving online dynamics estimation through richer sensing and system identification, extending dynamics-aware representations to articulated and deformable objects, and integrating DAPL with higher-level planning for long-horizon manipulation in complex environments. We believe dynamics-aware representations offer a promising direction for advancing dexterous manipulation beyond grasp-centric paradigms.

\section*{Acknowledgment}
This work was partially supported by New Generation Artificial Intelligence-National Science and Technology Major Project (No. 2025ZD0122905).

\bibliographystyle{plainnat}
\bibliography{main}

\clearpage
\appendix















\setcounter{equation}{0}
\setcounter{table}{0}
\setcounter{figure}{0}

\subsection{Details of World Model}

\subsubsection{Scene Representation Construction.}
This section provides additional implementation details for constructing the scene representation used by the world model. Our policy operates on a physical scene representation $\mathcal{P} \in \mathbb{R}^{N \times 7}$, synthesized from three distinct components: the target object, environmental obstacles, and the robot’s end-effector. To ensure a fixed-size input tensor while maintaining high geometric fidelity, we implement a hierarchical sampling and coordinate transformation pipeline. 

\textbf{Sampling and  Filtering.}
For each object in the scene, a canonical point set containing 512 points is pre-sampled from its original mesh using a uniform random sampling strategy. At runtime, canonical points are transformed to the world frame using the object pose $T \in SE(3)$. and its specific scaling factor. The end-effector is similarly represented by 256 points sampled from its corresponding mesh.
To handle cluttered environments containing a variable number of obstacles without exceeding computational constraints, we employ a selection mechanism. Instead of processing the entire background point cloud, we aggregate the points from all environmental obstacles and retain only the 512 points that are spatially closest to the centroid of the target object. This strategy prioritizes the most relevant local geometric constraints and contact boundaries, ensuring that the policy focuses on the immediate physical bottlenecks within the workspace.

\textbf{Physical Attribute.}
To empower the policy with the capability to reason about object dynamics and inertial effects, each point $\mathbf{p}_i$ in the aggregated scene point cloud is represented as a 7-dimensional feature vector:
\begin{equation}
    \mathbf{f}_i = [x, y, z, m, v_x, v_y, v_z]
\end{equation}
where $[x, y, z]$ denotes the Cartesian coordinates and $[v_x, v_y, v_z]$ captures the instantaneous linear velocity in the world frame. 
Under the rigid-body assumption, we embed physical properties directly into the point-wise features by distributing the object's total mass $M$ across its sampled points. Specifically, each point is assigned a scalar mass value:
\begin{equation}
    m = \frac{M}{N}
\end{equation}
where $N$ is the number of points sampled for that specific object. This \textit{dynamic-aware} encoding allows the network to implicitly distinguish between objects of different physical identities and reason about their kinetic states. Such a representation is crucial for predicting collision outcomes and handling complex manipulation dynamics involving varying mass distributions. During the training phase, all 7-dimensional features are normalized using running statistics to ensure numerical stability across different physical scales.

\subsubsection{Architecture}
Given observations at time $t$, the model predicts the scene state at $t+\Delta t$ ($\Delta t = 0.1$s). The model follows an encoder-decoder architecture:

\textbf{Tokenization.}
The input scene point cloud $\mathcal{P} \in \mathbb{R}^{1,280 \times 7}$ is first partitioned into $P=40$ patches using a semantic-aware sampling strategy. To maintain semantic integrity, we independently sample 16 patches from the target object, 16 from environmental obstacles, and 8 from the robot end-effector via Farthest Point Sampling (FPS) and $k$-Nearest Neighbors ($k$-NN, $K=32$). This prevents feature cross-contamination between the robot and its environment at the earliest stage.
Each patch is normalized to its local coordinate system and processed through a PointNet-style Patch Encoder—comprising a two-layer MLP followed by a max-pooling operation—to generate initial tokens. By integrating spatial coordinates with mass and velocity attributes, these tokens become \textit{dynamic- } before being projected into a $D$-dimensional latent space ($D=128$) and augmented with Sinusoidal Position Embeddings to provide spatial context.

\textbf{Dynamics Encoder.} The latent patch tokens are processed by a Vision Transformer (ViT) backbone, consisting of 12 blocks and 8 heads with a hidden dimension of 128. This stage extracts high-level spatial-temporal features $\mathbf{Z}_t$ that capture the global geometric context of the scene.

\textbf{Decoder.} To incorporate the robot's action, the end-effector flow between two frames, $\mathbf{a}_t \in \mathbb{R}^3$, is projected into a latent action token and fused with the patch tokens $\mathbf{Z}_t$ via a single cross-attention Transformer layer. The resulting action-conditioned tokens are then mapped back to the point-wise resolution through an unpatchify scatter operation. Specifically, we use the patch indices and patch centers preserved from the encoding stage to assign patch embeddings back to their constituent points. Patch features are scattered back to the point level using stored patch indices. When a point belongs to multiple patches, features are averaged. For each point within a patch, its 3D offset relative to the patch center is processed by a MLP to generate a local positional embedding. This encoding is summed with the corresponding patch embedding, yielding refined point-level features that integrate global semantic context with fine-grained local geometry.

\subsubsection{Learning Objective.}
The model employs parallel heads to predict the future state of the scene.
The Position Head directly regresses the per-point coordinates $\hat{P}_{t+\Delta t}$ for the next frame,
while the Velocity Head predicts the per-point future velocities $\hat{V}_{t+\Delta t}$.
In addition, we introduce a global velocity-variance matching term that aligns the per-dimension variance
of the predicted velocities with that of the ground-truth velocities, encouraging the model to capture
the overall distribution of motion in the scene.

We decompose the training objective into three components:
\begin{align}
\mathcal{L}_{\text{pos}} &= \big\| \hat{P}_{t+\Delta t} - P_{t+\Delta t} \big\|_2^2, \\
\mathcal{L}_{\text{vel}} &= \big\| \hat{V}_{t+\Delta t} - V_{t+\Delta t} \big\|_2^2, \\
\mathcal{L}_{\text{var}} &= \big\| \mathrm{Var}(\hat{V}_{t+\Delta t}) - \mathrm{Var}(V_{t+\Delta t}) \big\|_2^2,
\end{align}
where $\mathrm{Var}(\cdot)$ denotes the per-dimension empirical variance of the velocity over all points
and batch samples. The final loss is a weighted sum:
\begin{equation}
\mathcal{L}
= \lambda_{\text{pos}} \mathcal{L}_{\text{pos}}
+ \lambda_{\text{vel}} \mathcal{L}_{\text{vel}}
+ \lambda_{\text{var}} \mathcal{L}_{\text{var}},
\end{equation}
and we set $\lambda_{\text{pos}} = 1.0$, $\lambda_{\text{vel}} = 1.0$, and $\lambda_{\text{var}} = 100.0$ in all experiments.
The high weight on $\mathcal{L}_{\text{var}}$ is crucial to prevent the model from collapsing to trivial, near-zero velocity predictions and to maintain a realistic spread of motion.

\subsection{Details of Policy Learning}

\label{supp:policy_learning}
\begin{table*}[t]
    \centering
    \caption{Summary of our MDP components. $\dagger$: simulation only.}
    \label{tab:mdp_summary}
    \small
    \renewcommand{\arraystretch}{1.3}
    \begin{tabularx}{\textwidth}{|l|c|c|X|}
        \hline
        \rowcolor[gray]{0.9}
        \textbf{State Component} & \textbf{Symbol} & \textbf{Dimension} & \multicolumn{1}{c|}{\textbf{Description}} \\ \hline
        Object point cloud & $P_o$ & $\mathbb{R}^{512 \times 7}$ & Surface-sampled object point cloud in the environment frame, with per-point mass and contact-conditioned velocity $(x,y,z,m,v_x,v_y,v_z)$ \\ \hline
        Environment point cloud & $P_e$ & $\mathbb{R}^{512 \times 7}$ & Surface-sampled obstacle point cloud in the environment frame, with per-point mass and velocity $(x,y,z,m,v_x,v_y,v_z)$ \\ \hline
        End-effector point cloud & $P_{ee}$ & $\mathbb{R}^{256 \times 7}$ & Surface-sampled end-effector point cloud in the environment frame, with per-point mass and velocity $(x,y,z,m,v_x,v_y,v_z)$ \\ \hline
        Hand state & $s_t^{EE}$ & $\mathbb{R}^9$ & End-effector position and orientation in 6D rotation representation (3D position + 6D orientation) \\ \hline
        Robot state & $s_t^q$ & $\mathbb{R}^{14}$ & Joint positions and velocities of the 7-DoF arm \\ \hline
        Relative goal pose & $T_g$ & $\mathbb{R}^9$ & Target object pose expressed relative to the current object pose (3D translation + 6D orientation) \\ \hline
        Physics parameters$^\dagger$ & $\rho$ & $\mathbb{R}^5$ & Object mass, object static friction, hand friction, ground friction, and object restitution \\ \hline
        Previous action & $a_{t-1}$ & $\mathbb{R}^7$ & Previous-step joint-space residual action applied to the arm \\ \hline \hline
        \textbf{Action Component} & \textbf{Symbol} & \textbf{Dimension} & \multicolumn{1}{c|}{\textbf{Description}} \\ \hline
        Joint-space residuals & $\Delta q_t$ & $\mathbb{R}^7$ & Relative joint position commands: $q^{\text{target}} = q_t + \Delta q_t$ for the 7 arm joints \\ \hline
    \end{tabularx}
    \label{tab:mdp_observation}
\end{table*}

\subsubsection{Architecture}
The policy architecture is a unified pipeline that maps multi-modal observations, as detailed in Table \ref{tab:mdp_observation}, to actions through a hierarchical fusion process. 
First, the dynamics encoder processes scene point clouds to extract latent patch tokens $\mathbf{Z} \in \mathbb{R}^{P \times D}$. Simultaneously, an Environment State $\mathbf{s}_{env}$ is formed by aggregating the task goal, robot joint states, physical parameters, and the last action. To integrate these modalities, we adopt a cross-attention layer following CORN, where $\mathbf{s}_{env}$ is projected into a query token to perform cross-attention over $\mathbf{Z}$. The resulting task-conditioned feature $\mathbf{z}_{attn}$ is then concatenated with $\mathbf{s}_{env}$ and passed through a fusion MLP(hidden layers: $[512, 256, 128]$) to yield a compact global representation $\mathbf{g}$.
Finally, $\mathbf{g}$ is fed into the Actor and Critic heads, both implemented as two-layer MLPs with a hidden dimension of $[64]$. The Actor head outputs the mean of a Gaussian distribution with a learnable $\log\_std$, while the Critic head provides the state-value estimate.

\subsubsection{Training and Hyperparameters}
The policy is trained within the Isaac Lab simulator using the rsl-rl implementation of the PPO algorithm. To preserve the physical priors captured during pre-training, the Dynamics Encoder is loaded with pre-trained weights and remains frozen throughout the reinforcement learning process. Training is accelerated using 2048 parallel environments on an $8 \times$ L40 GPU cluster, providing a high throughput of diverse contact interactions and trajectory samples. 
The specific hyperparameter configuration for the PPO algorithm is detailed in Table \ref{tab:ppo_hyperparams}.

\begin{table}[h]
\centering
\caption{PPO Algorithm Hyperparameters}
\label{tab:ppo_hyperparams}
\begin{tabular}{lc}
\hline
\textbf{Parameter} & \textbf{Value} \\ \hline
Value loss coefficient & 0.5 \\
Use clipped value loss & True \\
Clip parameter ($\epsilon$) & 0.3 \\
Entropy coefficient & 0.006 \\
Learning epochs & 8 \\
Mini-batches & 8 \\
Learning rate & $5.0 \times 10^{-5}$ \\
LR Schedule & Adaptive \\
Discount factor ($\gamma$) & 0.99 \\
GAE parameter ($\lambda$) & 0.95 \\
Desired KL divergence & 0.016 \\
Max gradient norm & 1.0 \\ \hline
\end{tabular}
\end{table}

\subsubsection{Reward Function}
\label{sec:supp_reward}

\textbf{Notation.}
Let $p_{\text{obj}} \in \mathbb{R}^3$ denote the position of the target object and
$p_{\text{des}} \in \mathbb{R}^3$ denote the desired target position.
The object orientation and target orientation are represented as unit quaternions
$q_{\text{obj}}$ and $q_{\text{des}}$, respectively.
The left and right end-effector positions are denoted as
$p_{\text{ee,L}}, p_{\text{ee,R}} \in \mathbb{R}^3$.

\textbf{End-effector Proximity Reward.}
We encourage physical interaction by rewarding proximity between the end-effector
and the target object.
The minimum distance between the object and either end-effector is computed as
\begin{equation}
d_{\text{ee}} =
\min\left(
\|p_{\text{obj}} - p_{\text{ee,L}}\|,
\|p_{\text{obj}} - p_{\text{ee,R}}\|
\right).
\end{equation}

The contact reward is defined using a smooth $\tanh$ shaping:
\begin{equation}
r_{\text{contact}} =
1 - \tanh\left(\frac{d_{\text{ee}}}{\sigma_{\text{contact}}}\right),
\end{equation}
where $\sigma_{\text{contact}} = 0.1$.

\textbf{Gated Object Goal Tracking Reward.}
Goal-related rewards are activated only when the end-effector is sufficiently close
to the target object:
\begin{equation}
\mathbb{I}_{\text{near}} =
\begin{cases}
1, & d_{\text{ee}} < d_{\text{th}}, \\
0, & \text{otherwise},
\end{cases}
\end{equation}
where $d_{\text{th}} = 0.1$\,m.
The position error is defined as
\begin{equation}
d_p = \|p_{\text{des}} - p_{\text{obj}}\|.
\end{equation}

The orientation error is computed using the quaternion angular distance:
\begin{equation}
d_r = 2 \arccos\left(\left| q_{\text{obj}} \cdot q_{\text{des}} \right|\right),
\quad d_r \in [0, \pi].
\end{equation}

The combined pose error is defined as
\begin{equation}
d = d_p + \frac{d_r}{5}.
\end{equation}

\textbf{Coarse Goal Tracking Reward.}
The coarse goal tracking reward is defined as
\begin{equation}
r_{\text{goal}} =
\mathbb{I}_{\text{near}}
\left(
1 - \tanh\left(\frac{d}{\sigma_{\text{coarse}}}\right)
\right),
\end{equation}
where $\sigma_{\text{coarse}} = 0.6$.

\textbf{Fine-grained Goal Tracking Reward.}
To encourage precise pose alignment, we introduce a fine-grained variant:
\begin{equation}
r_{\text{goal-fine}} =
\mathbb{I}_{\text{near}}
\left(
1 - \tanh\left(\frac{d}{\sigma_{\text{fine}}}\right)
\right),
\end{equation}
where $\sigma_{\text{fine}} = 0.3$.

\textbf{Task Success Reward with Clutter Disturbance Penalty.}
Position success is evaluated on the planar distance:
\begin{equation}
d_{\text{pos}} =
\left\|
p_{\text{des}}^{xy} - p_{\text{obj}}^{xy}
\right\|.
\end{equation}

Rotation success is defined as
\begin{equation}
d_{\text{rot}} =
2 \arccos\left(
\left| q_{\text{obj}} \cdot q_{\text{des}} \right|
\right).
\end{equation}

The success mask is defined as
\begin{equation}
m =
\mathbb{I}(d_{\text{pos}} < \tau_p)
\land
\mathbb{I}(d_{\text{rot}} < \tau_r).
\end{equation}

The base success reward is
\begin{equation}
r_0 =
\begin{cases}
1.0, & m = 1, \\
0, & m = 0.
\end{cases}
\end{equation}

\textbf{Obstacle Motion Penalty.}
Let $N_{\text{obs}}$ denote the number of non-target objects.
The average displacement and rotation are computed as
\begin{equation}
\bar{d} =
\frac{1}{N_{\text{obs}}}
\sum_{i=1}^{N_{\text{obs}}} d_{\text{obs}, i},
\quad
\bar{\theta} =
\frac{1}{N_{\text{obs}}}
\sum_{i=1}^{N_{\text{obs}}} \theta_{\text{obs}, i}.
\end{equation}

The normalized motion measures are defined as
\begin{equation}
\hat{d} =
\text{clip}\left(
\frac{\bar{d}}{d_{\max}}, 0, 1
\right),
\quad
\hat{\theta} =
\text{clip}\left(
\frac{\bar{\theta}}{\theta_{\max}}, 0, 1
\right).
\end{equation}

The combined motion score is
\begin{equation}
m_{\text{motion}} =
\frac{\hat{d} + \hat{\theta}}{2}.
\end{equation}

The scaling factor applied to the success reward is
\begin{equation}
s =
\text{clip}\left(
1 - 0.5 \cdot m_{\text{motion}},
0.5,
1.0
\right).
\end{equation}

The final success reward is defined as
\begin{equation}
r_{\text{success}} = r_0 \cdot s.
\end{equation}

\textbf{Total Reward.}
The total reward is computed as a weighted sum of all reward components:
\begin{equation}
R_{\text{total}} = \sum_i w_i r_i.
\end{equation}

\begin{table}[h]
\centering
\caption{Reward terms and weights used for RL training.}
\begin{tabular}{l c}
\hline
Reward term & Weight \\
\hline
$r_{\text{contact}}$ & 1.0 \\
$r_{\text{goal}}$ & 5.0 \\
$r_{\text{goal-fine}}$ & 16.0 \\
$r_{\text{success}}$ & 2000.0 \\
$\sigma_{\text{coarse}}$ & 0.6 \\
$\sigma_{\text{fine}}$ & 0.3 \\
\hline
\end{tabular}
\end{table}

\begin{figure*}[htbp]
  \centering
  \includegraphics[width=0.8\linewidth]{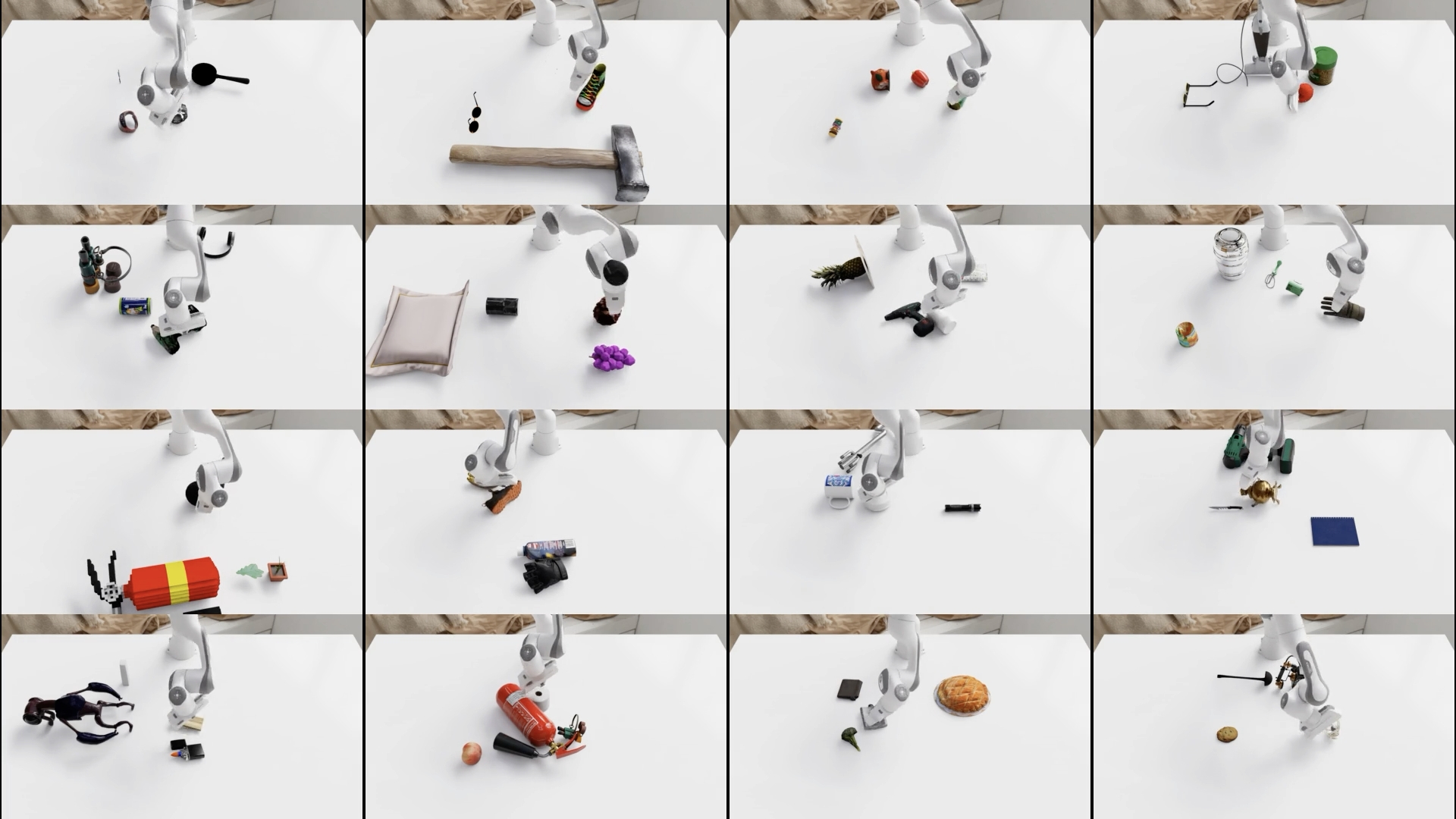}
  \caption{Visualizations of representative scenarios within the \textit{Sparse} clutter level.}
  \label{fig:scene_sparse}
\end{figure*}

\begin{figure*}[htbp]
  \centering
  \includegraphics[width=0.8\linewidth]{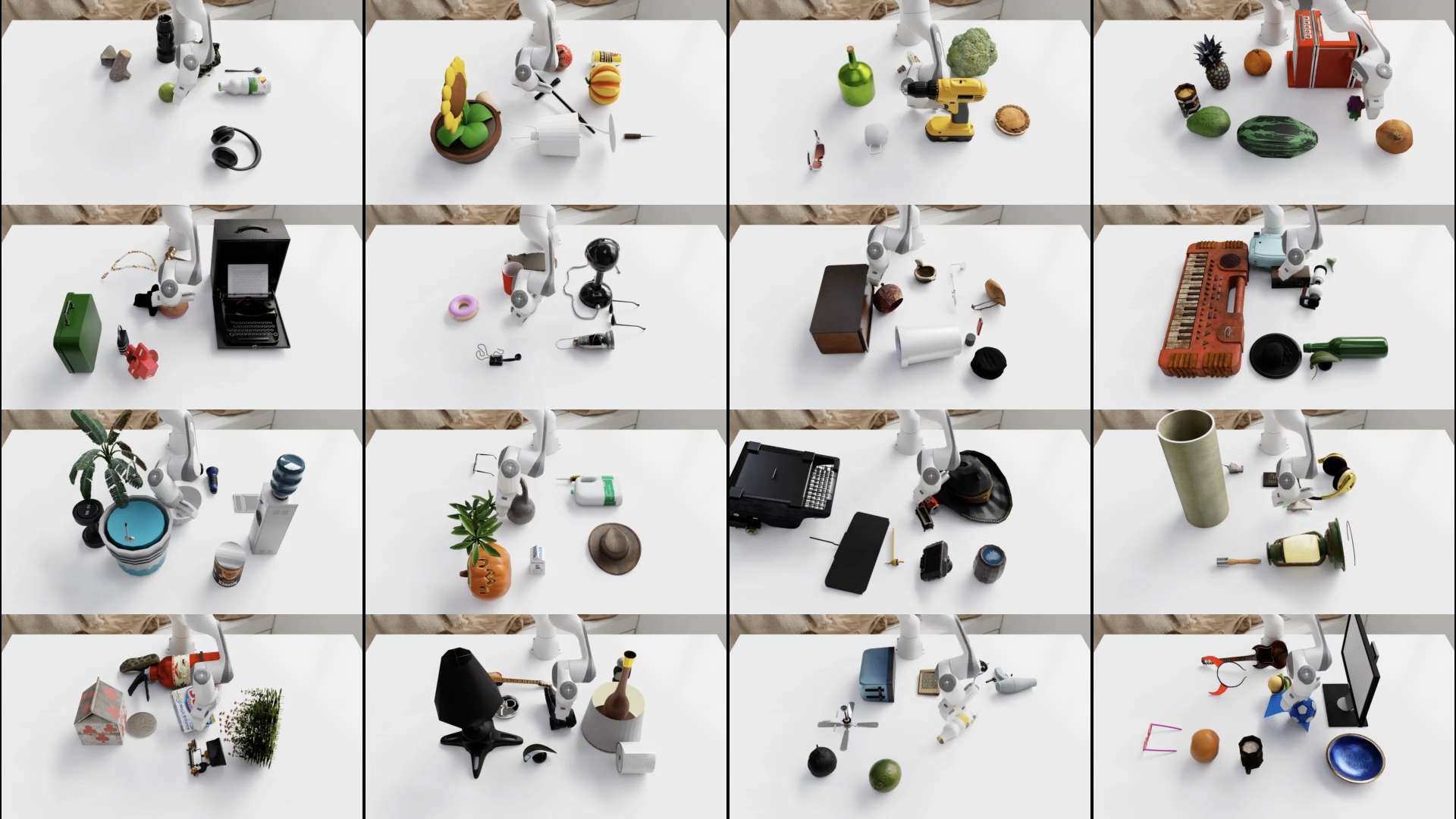}
  \caption{Visualizations of representative scenarios within the \textit{Moderate} clutter level.}
  \label{fig:scene_moderate}
\end{figure*}

\begin{figure*}[htbp]
  \centering
  \includegraphics[width=0.8\linewidth]{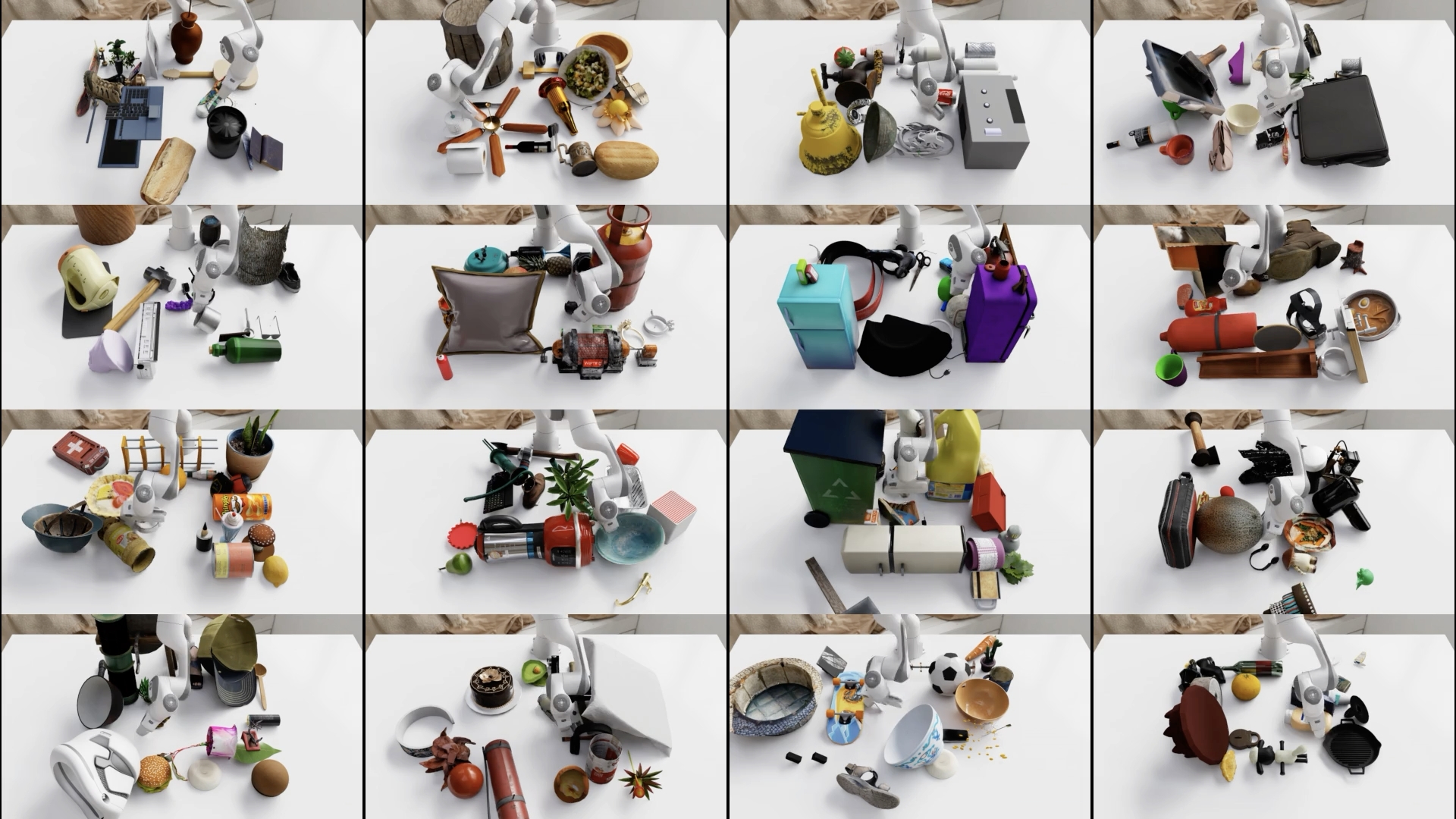}
  \caption{Visualizations of representative scenarios within the \textit{Dense} clutter level.}
  \label{fig:scene_dense}
\end{figure*}

\subsection{Details of Simulation Experiments}
\subsubsection{Simulation Scene Generation}

To ensure high diversity and physical plausibility, we extend the programmatic scene generation framework from GenManip~\cite{gao2025genmanip} within NVIDIA Isaac Sim. Our pipeline is specifically tailored for non-prehensile manipulation by procedurally scaling environmental complexity. Specifically, we define three difficulty levels—\textit{Sparse, Moderate, Dense}—determined by the density and spatial arrangement of obstacles. Representative scenes illustrating these varying levels of environmental clutter and task difficulty are presented in Fig.~\ref{fig:scene_dense},Fig.~\ref{fig:scene_moderate},Fig.~\ref{fig:scene_sparse}.

\textbf{Object Selection and Geometry Filtering.}
We curate our object library from the Objaverse dataset~\cite{deitke2023objaverse}, normalizing each object to a unit cube ($1\,\text{m}^3$) for numerical stability in the physical solver. To ensure real-time simulation performance and curriculum-based scene generation, we exclude objects exceeding $50,000$ vertices or $5\,\text{MB}$ in size, and further categorize the remaining assets into \textit{large} (scale $\geq 0.25$) and \textit{small} (scale $< 0.25$) cohorts to facilitate strategic placement of targets in environmental clutter.

\textbf{Physically Grounded Initialization.}
A primary challenge in procedural generation is ensuring objects are initialized in physically stable poses to avoid jitter or explosive contact at the start of a simulation. We utilize the trimesh library to pre-compute a set of stable candidate orientations for each object based on its convex hull. For each object $i$, the initial pose is defined as $P_{i,\text{init}} = (\mathbf{p}_i, \mathbf{q}_i)$, where the orientation $\mathbf{q}_i$ is randomly sampled from a pre-calculated stable set $\mathcal{Q}_{i,\text{stable}}$ (comprising up to 64 candidates). This sampling strategy ensures high variance in task configurations while maintaining physical realism.

\textbf{Scene Composition and Clutter Management.}
Our environment supports four difficulty levels, defined by the density and size of obstacles, as summarized in Table~\ref{tab:difficulty_levels}.

\begin{table}[h]
\centering
\caption{Difficulty levels and procedural object counts.}
\label{tab:difficulty_levels}
\begin{tabular}{@{}lccc@{}}
\toprule
Difficulty & Large Obstacles & Small Obstacles & Target \\ \midrule
Sparse       & 1               & 2               & 1              \\
Moderate     & 3               & 4               & 1              \\
Dense  & 5              & 6               & 1              \\ \bottomrule
\end{tabular}
\end{table}

Objects are placed using a recursive search for non-overlapping "on-table" relations. To ensure reachability, the target object's initial position is constrained within a central region defined by $x \in [-0.15, 0.15]$ and $y \in [-0.3, 0.3]$ meters relative to the table center.

\textbf{Goal Pose Generation.}
For each generated scene configuration, we define a suite of manipulation tasks by generating 16 distinct pairs of initial and goal poses $(P_{\text{target, init}}, P_{\text{target, goal}})$ for target object. To ensure a non-trivial manipulation task, each goal pose must satisfy a minimum planar displacement constraint:
\begin{equation}
||\mathbf{p}_{\text{target, goal}}^{2D} - \mathbf{p}_{\text{target, init}}^{2D}||_2 \geq 0.15\,\text{m}
\end{equation}
where $\mathbf{p}^{2D}$ denotes the coordinates on the tabletop surface. Both the initial and goal poses are sampled from the pre-computed physically stable orientation set $\mathcal{Q}_{\text{stable}}$.

\begin{figure*}[htbp]
  \centering
  \includegraphics[width=0.9\linewidth]{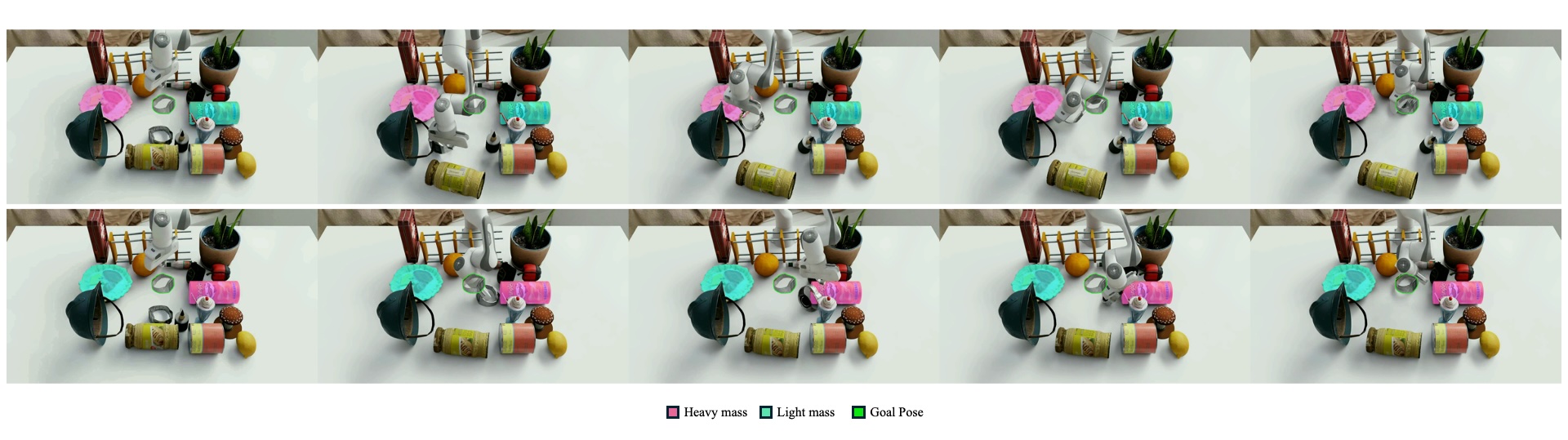}
  \caption{\textbf{Adaptive Interaction under Mass Perturbations.} The policy reconfigures trajectories by treating heavy objects as stable anchors. \textbf{Case A:} Utilizing the heavy pie for support. \textbf{Case B:} Leveraging the heavy can as a mechanical pivot after mass swapping.}
  \label{fig:Adaptive behavior}
\end{figure*}

\subsubsection{Details of Baselines.}

To evaluate the effectiveness of our proposed framework, we compare it against several representative baselines across different categories:

\begin{itemize}
    \item \textbf{Prehensile Manipulation (GraspGen + CuRobo):} This baseline represents a traditional decoupled approach. We use \textbf{GraspGen}~\cite{murali2025graspgen} to generate diverse candidate grasp poses for the target object from point cloud observations. Subsequently, \textbf{CuRobo}~\cite{sundaralingam2023curobo}, a high-performance GPU-accelerated motion planner, is employed to find collision-free trajectories to the target. This baseline evaluates the limits of traditional grasping in cluttered environments where non-prehensile strategies might be more efficient.
    
    \item \textbf{Human Teleoperation:} To establish an upper-bound performance metric, we collect expert demonstrations using the \textbf{GELLO}~\cite{wu2024gello} teleoperation interface. For evaluation, we randomly select 10 distinct scenes from each of the three difficulty levels (\textit{Sparse, Moderate,} and \textit{Dense}), totaling 30 unique test configurations. Unlike automated baselines restricted to specific modalities, the human operator is permitted to employ \textbf{both prehensile and non-prehensile strategies} fluidly.
    
    \item \textbf{Representation Learning + RL:} We benchmark against several state-of-the-art representation learning methods combined with Reinforcement Learning (RL):
    \begin{itemize}
        \item \textit{General Encoders:} We utilize \textbf{Point2Vec}~\cite{abouzeid2023point2vec} and \textbf{Concerto}~\cite{zhang2025concerto} as foundational point cloud backbones. These models are pre-trained on large-scale 3D datasets and are frozen or fine-tuned to extract task-relevant geometric features for the RL policy. Notably, as the Point Transformer V3 architecture used in \textbf{Concerto} poses significant computational challenges for inference during large-scale RL training, we employ its \textbf{Tiny variant} to ensure experimental feasibility and efficiency.
        \item \textit{Non-prehensile Tailored Encoders:} We compare against \textbf{CORN}~\cite{cho2024corn} and \textbf{UniCORN}~\cite{cho2025hierarchical}, which are designed to capture contact-centric features. Since the original CORN focuses on single-object tasks, we implement \textbf{CORN-multi}: it independently encodes each object's point cloud and integrates their features via the State-Dependent Cross-Feature Network mentioned in \ref{supp:policy_learning} to represent the multi-object scene. 
    \end{itemize}
    To ensure a fair comparison, all baseline methods share the same policy and value network architectures as our approach, with the only difference being the choice of the point cloud encoder. Furthermore, since these baselines do not involve pre-training on physical attributes or velocity information, they exclusively take 3D coordinates $(x, y, z)$ as input. This controlled setup allows us to isolate the impact of our dynamics-aware representation learning from other architectural factors.
\end{itemize}

\subsubsection{More Analysis on Adaptive Behavior.}
Beyond quantitative performance gains, our policy exhibits adaptive interaction strategies that reflect a deep understanding of contact-induced dynamics. To rigorously evaluate this, we conduct a \textit{controlled mass-perturbation experiment} using a scene containing a pie and a Pringles can as potential interactors. We create two contrasting settings by scaling the default mass of these objects by factors of 100$\times$ (Heavy) and 0.1$\times$ (Light).

As illustrated in Fig.\ref{fig:Adaptive behavior}, the resulting trajectories under identical initial and goal configurations are markedly different:
\begin{itemize}
    \item \textbf{Case A (Heavy Pie, Light Can):} When the pie is assigned a heavy mass, the policy proactively exploits it as a \textit{stable mechanical anchor}. It establishes deliberate contact with the pie to generate the necessary reaction forces for reorienting the target, while simultaneously executing a wider swing to avoid the lightweight Pringles can, which would otherwise be easily displaced.
    \item \textbf{Case B (Light Pie, Heavy Can):} Conversely, when the mass assignments are swapped, the policy autonomously adapts its path. It now treats the Pringles can as the functional support, utilizing its high inertia to stabilize the manipulation maneuver while carefully bypassing the now-unstable pie to prevent unintended scene destabilization.
\end{itemize}

\begin{figure*}[htbp]
  \centering
  \includegraphics[width=0.9\linewidth]{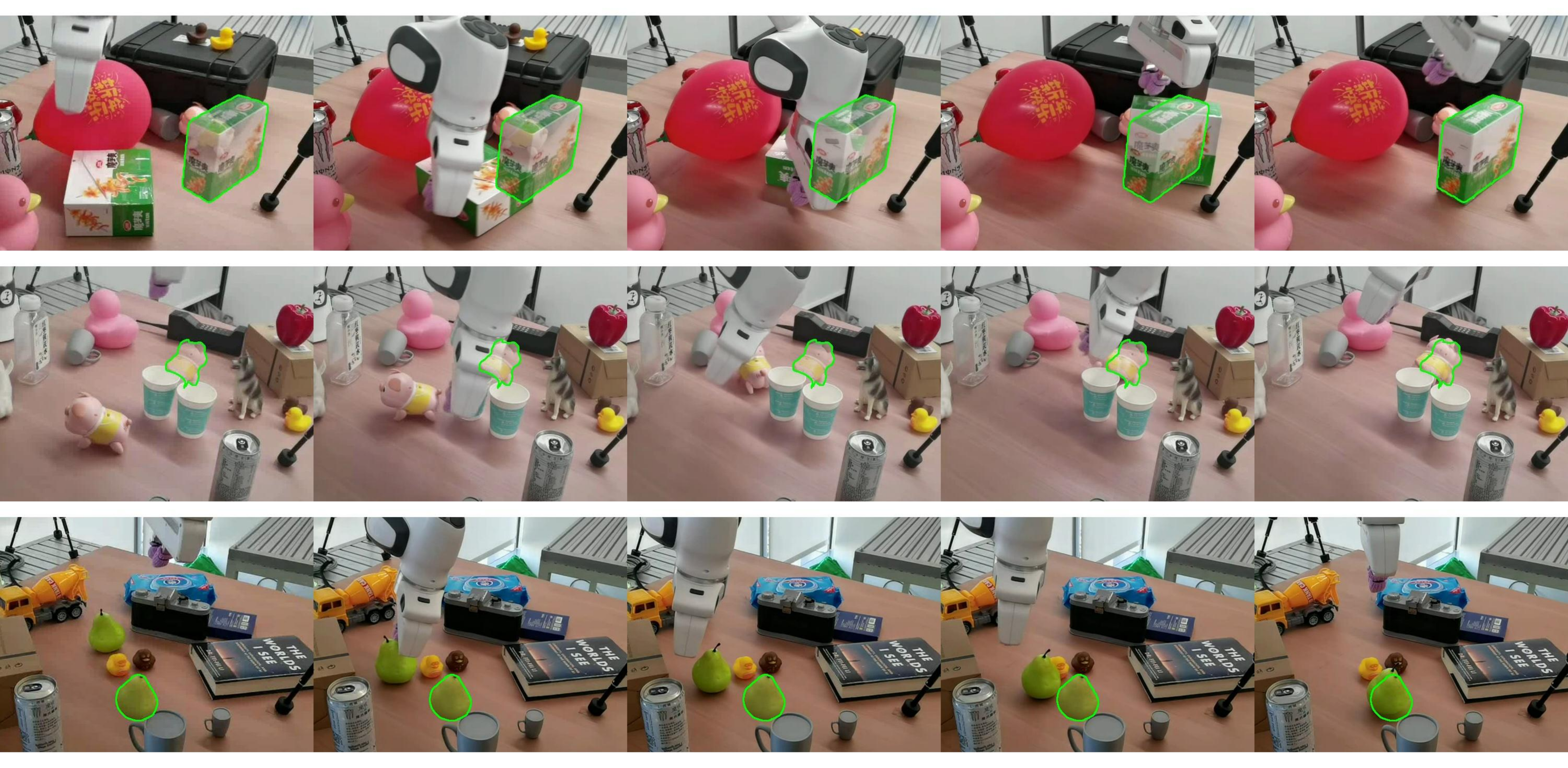}
  \caption{\textbf{Disturbance Avoidance:} The policy bypasses lightweight objects to maintain scene stability.}
  \label{fig:real_1}
\end{figure*}

\begin{figure*}[htbp]
  \centering
  \includegraphics[width=0.9\linewidth]{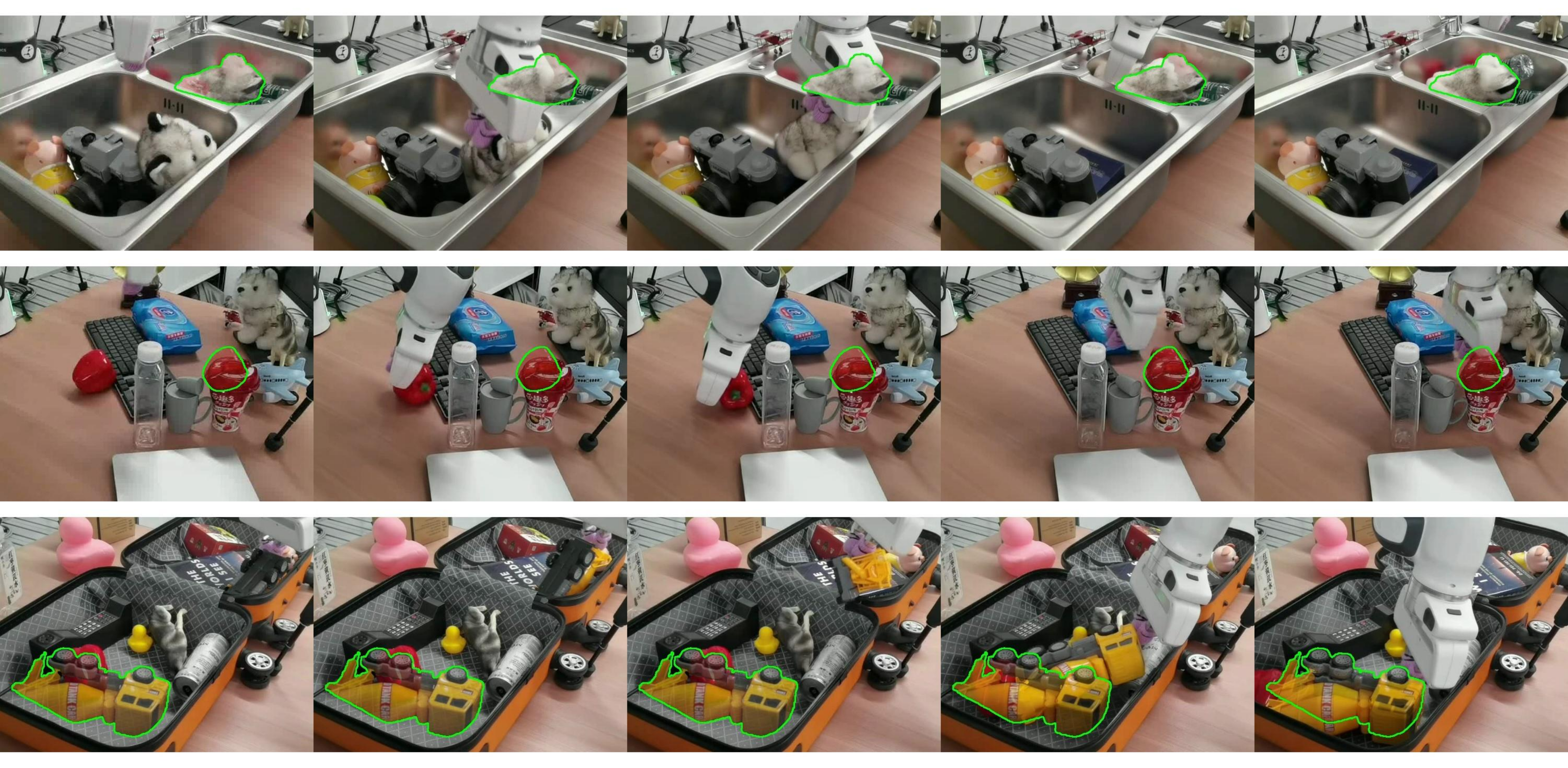}
  \caption{\textbf{Obstacle Traversal:} The policy manipulates the target to traverse physical barriers (e.g., sink dividers).}
  \label{fig:real_2}
\end{figure*}

\begin{figure*}[htbp]
  \centering
  \includegraphics[width=0.9\linewidth]{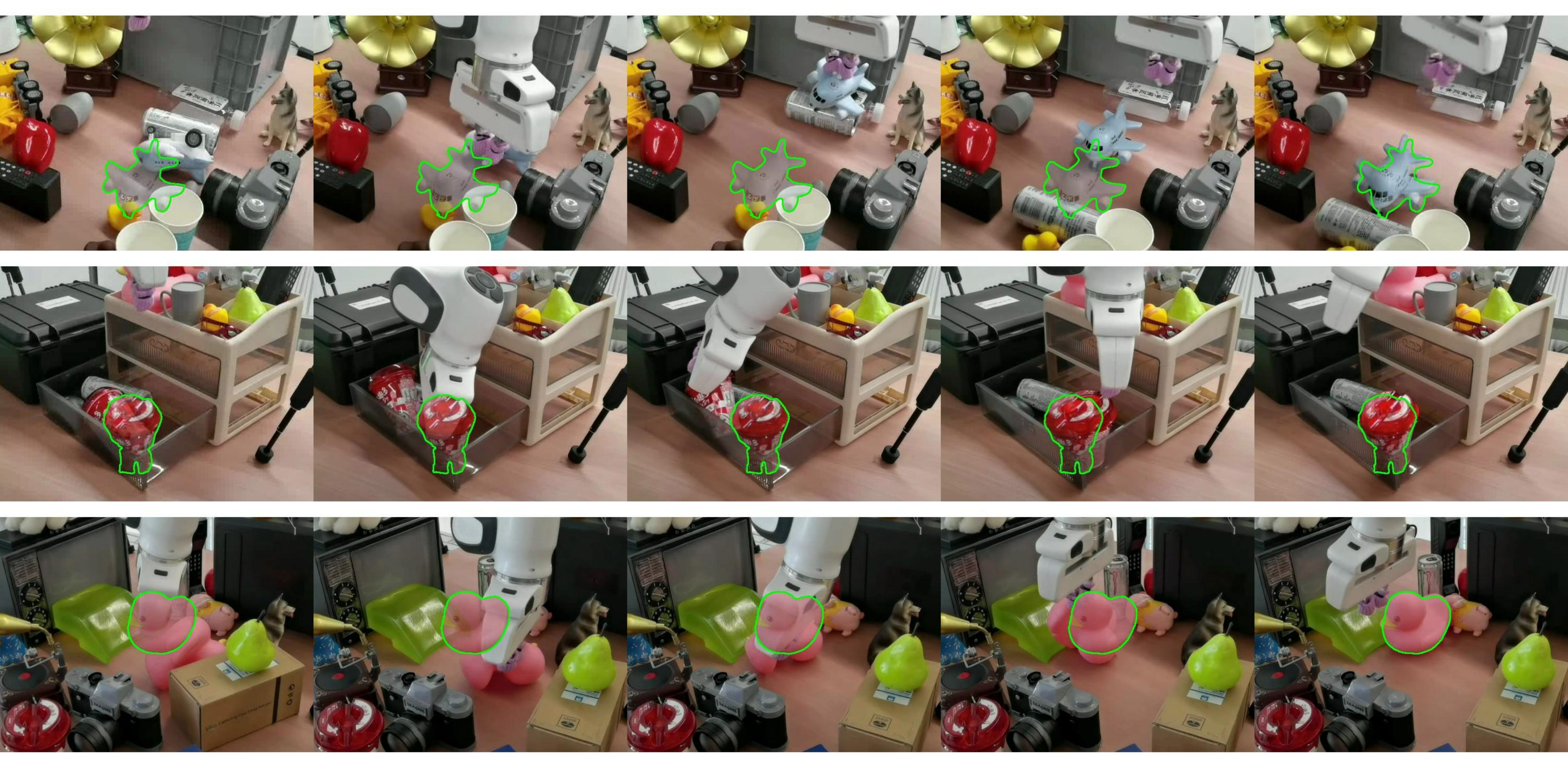}
  \caption{\textbf{Leveraging Contact:} The policy utilizes stable environmental objects as anchors to facilitate reorientation.}
  \label{fig:real_3}
\end{figure*}

These results provide strong evidence that our dynamics-aware representation allows the policy to reason about the \textit{causal consequences} of interaction. Rather than relying on static geometric heuristics, the policy decodes latent physical properties from the point cloud sequences to determine which environmental objects can be leveraged and which must be avoided to ensure task success and scene integrity.

\subsection{Real-world Experiment and Sim-to-Real Deployment}

\subsubsection{System Configuration}
We deploy our policy on a Franka Research 3robot arm equipped with three Intel RealSense cameras for comprehensive multi-view coverage. Our perception pipeline integrates SAM2 for initial frame segmentation, XMem for online tracking, and FoundationPose for continuous 6D pose estimation. To provide physical priors, object mass is estimated via GPT-5 using a zero-shot prompt: \textit{``You are an expert in physical property estimation... estimate its mass in kilograms.''} This enables the robot to reason about dynamics without specialized force sensors.

\subsubsection{Dynamics Distillation and Estimation}
A primary obstacle in Sim-to-Real transfer is the significant discrepancy between simulation and reality regarding physical states. First, the real-world policy lacks access to \textbf{privileged information} (e.g., ground-truth mass and friction). Second, unlike the precise states in simulation, real-world observations—including \textbf{object positions, velocities, and estimated masses}—are inherently noisy and prone to estimation errors. We bridge this gap through an integrated distillation framework:

\begin{itemize}
    \item \textbf{Teacher-Student Distillation under Perturbation:} We employ a teacher-student scheme to internalize dynamics reasoning. During training, the teacher policy receives noise-free, privileged information to learn optimal strategies. Simultaneously, the student, is trained to replicate the Teacher's latent representation and actions. Crucially, we inject significant Gaussian noise ($\mu=0, \sigma=0.05$) into the Student's input observations (coordinates and velocities). 
    \item \textbf{Inferring Effective Dynamics:} This ``distillation under perturbation'' forces the dynamics-aware encoder to filter sensory noise and infer underlying ``effective dynamics'' from temporal cues. Consequently, the policy becomes resilient to the coarse mass estimates from \textbf{GPT-5} and the jittery velocities derived from \textbf{FoundationPose} delta poses.
    \item \textbf{EKF-Refined Velocity:} To further support the policy during deployment, real-world velocities are derived from consecutive pose differences and refined via an \textbf{Extended Kalman Filter (EKF)} to provide a cleaner temporal signal matching the Student's learned distribution.
\end{itemize}

\subsubsection{Safe and Conservative Action Mapping}
Direct deployment often fails due to hardware torque limits and excessive movements. Following HAMNet, we regulate robot motion through two strategies:

\textbf{Action Magnitude Curriculum.} To encourage safe trajectories, we implement a decaying schedule for the \textbf{action scale} of the subgoal residuals ($\Delta q$). During training, this scale is gradually reduced from \textbf{0.1 to 0.01}. This curriculum guides the policy from coarse global exploration to fine-grained adjustments, preventing over-shooting and high-torque impacts.

\textbf{Cartesian-space Action Clipping.} We implement a Cartesian-space action clipping algorithm to prevent high-force impacts. We project joint-space residuals $\Delta q$ to Cartesian movement $\Delta x$ via the Jacobian $J$. If $\Delta x$ exceeds a bound $\epsilon_x$, the excess is mapped back to joint space using a damped pseudo-inverse Jacobian:
\begin{equation}
    \Delta q_{excess} = J^T(JJ^T + \lambda^2 I)^{-1}\Delta x_{excess}
\end{equation}
The final action, $\Delta q_{clamped} = \Delta q - \Delta q_{excess}$, is executed by an impedance controller to reach the desired positions smoothly while maintaining compliant contact.

\subsubsection{Experimental Protocol}
\textbf{Success Evaluation.}
We evaluate performance on 10 diverse real-world cluttered scenes against \textbf{FACTR}. A trial is successful if the target reaches the goal (recorded via \textbf{FoundationPose}) within a translational threshold of \textbf{0.05\,m} and a rotational threshold of \textbf{0.1\,rad} within \textbf{90 seconds}. 

\textbf{Symmetry-Aware Metric.}
For symmetric objects where orientation is ambiguous, rotational errors are computed only on the non-symmetric axes to ensure the metric reflects functional task completion.

\textbf{More Qualitative results.}
Additional qualitative results are presented in Fig.~\ref{fig:real_1},Fig.~\ref{fig:real_2},Fig.~\ref{fig:real_3}.



\end{document}